\documentclass[times,twocolumn,final]{elsarticle}

\usepackage{scs}
\usepackage{framed,multirow}

\usepackage{amssymb}
\usepackage{latexsym}
\usepackage{xspace}

\usepackage{url}
\usepackage{xcolor}
\usepackage{enumitem}
\usepackage{comment}
\RequirePackage[normalem]{ulem}
\usepackage{hyperref}
\usepackage{amsmath}
\usepackage[switch,pagewise]{lineno} 
\newcommand{\hidecomment}[1]{}

\newcommand{\highlight}[1]{{\color{blue}{#1}}}
\newcommand{\removed}[1]{{\color{red}\sout{#1}}}

\renewcommand{\highlight}[1]{#1}
\renewcommand{\removed}[1]{{\iffalse{#1}\fi}}

\journal{Sustainable Cities and Society}

\begin{document}

\verso{Pre-print version}

\begin{frontmatter}

\title{CitySurfaces: City-Scale Semantic Segmentation of Sidewalk Materials}%

\author[1,2]{Maryam Hosseini}
\ead{maryam.hosseini@nyu.edu}
\author[3]{Fabio Miranda}
\author[1]{Jianzhe Lin}
\author[1]{Claudio Silva}

\address[1]{Department of Computer Science and Engineering, New York University, NY, US}
\address[2]{Urban Systems, Rutgers University, NJ, US}
\address[3]{Department of Computer Science, University of Illinois at Chicago, IL, US}


    


\begin{abstract}
While designing sustainable and resilient urban built environment is increasingly promoted around the world, significant data gaps have made research on pressing sustainability issues challenging to carry out. Pavements are known to have strong economic and environmental impacts; however, most cities lack a spatial catalog of their surfaces due to the cost-prohibitive and time-consuming nature of data collection. 
Recent advancements in computer vision, together with the availability of street-level images, provide new opportunities for cities to extract large-scale built environment data with lower implementation costs and higher accuracy.
In this paper, we propose CitySurfaces, an active learning-based framework that leverages computer vision techniques for classifying sidewalk materials using widely available street-level images.
We trained the framework on images from New York City and Boston and the evaluation results show a $90.5\%$ mIoU score.
Furthermore, we evaluated the framework using images from six different cities, demonstrating that it can be applied to regions with distinct urban fabrics, even outside the domain of the training data.
CitySurfaces can provide researchers and city agencies with a low-cost, accurate, and extensible method to collect sidewalk material data which plays a critical role in addressing major sustainability issues, including climate change and surface water management. 

\end{abstract}


\end{frontmatter}
\section{Introduction}

\removed{The surface of cities has gone through tremendous changes since the formation of the early settlements. From stone causeways of ancient Egypt to crosswalks paved with wood planks, flagstones, granite or bricks~\citep{tillson1900street}; from natural soil to impervious concretes and dark, heat-absorbing asphalts of the industrialized cities~\citep{lay2020paving}, to modern permeable materials designed to address water runoff and excessive heat absorption challenges; city pavements evolved through time, responding to the new requirements of each era. However, in the race to conquer more land, the natural environment is put at risk, apart from the economic, social, and public health burdens created by uninformed decisions of the past.}

\highlight{As urban areas expand around the world, more impervious surfaces replace the natural landscape, creating significant ecological, hydrological, and economic disruptions~\citep{arnold1996impervious, chithra2015impacts}. Choosing the right material to cover city surfaces has become a critical issue in mitigating the adverse effects of increased anthropogenic activities.} 
Historically, local availability, cost, strength, and aesthetics were the main factors influencing the choice of \highlight{surface} pavements \citep{tillson1900street,lay2020paving}. The advent of asphalt and, later, concrete changed the face of cities. The longevity and durability coupled with relatively low production and installation costs made them the pavements of choice. However, as it was later revealed, these benefits came with huge environmental burdens~\citep{van2015towards}. \removed{The excessive use of impervious surfaces is shown to be the primary cause of the Combined Sewer Overflows (CSOs), which can lead to massive pollution of natural bodies of water and street flooding~\citep{joshi2021not}. Sidewalk pavements can also create public health hazards such as outdoor falls or pose a barrier to walkability and accessibility of public spaces, specifically for the more vulnerable population and wheelchair users~\citep{talbot2005falls, chippendale2015, thomas2020keep, CLIFTON200795,AGHAABBASI2018475}.}

\begin{figure*}
\centering
\includegraphics[width=\linewidth]{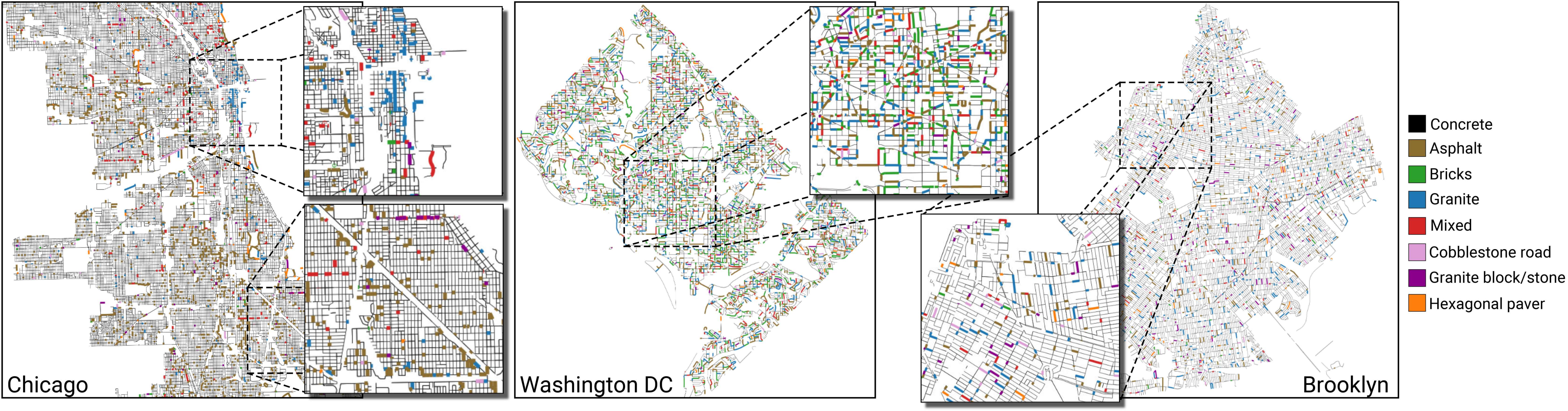}
\caption{\removed{Using CitySurfaces to classify paving materials from street-level images of Chicago, Washington DC, and Brooklyn (not part of our training data).
In each city, we collected street-level images at every five meters, covering the whole street network. The maps depict dominant paving materials in each
street segment. Segments where the dominant material differs from concrete are drawn using a thicker line. As can be seen, Washington DC uses
more diverse paving materials compared to Chicago and Brooklyn, where concrete is more widely used. However, downtown Chicago (top left) stands out
from the rest of the city in the way diverse materials such as granite paved the surface of the city.} \highlight{Using CitySurfaces to map the dominant surface material in Chicago, Washington DC, and Brooklyn (not part of our training data). Segments where the dominant material differs from concrete are drawn using a thicker line.} 
}
\label{fig:overview}
\end{figure*}

\highlight{One of the concerning environmental impacts of impervious surfaces is the sharp rise in urban temperature compared to its neighboring rural areas -- a phenomenon called Urban Heat Island (UHI) effect~\removed{\citep{mohajerani2017urban}}~\citep{oke1982energetic}. UHI, which poses serious challenges to public health, ecological environment, and urban liveability~\citep{estoque2017effects}, is shown to be directly associated with surface characteristics, such as thermal performance and reflectivity. It can influence microclimates within the city by absorbing more diurnal heat and emitting that into the atmosphere at night~\citep{polacco_study_2012, wu2018characterizing, nwakaire2020urban}. Natural surfaces and vegetation increase the amount of evapotranspiration and decrease the overall temperature and create a cool island effect~\citep{amati2010green, du2017quantifying}. Reflective/high-albedo materials are also known to decrease UHI~\citep{akbari2009global, santamouris2011using, santamouris2013using, zhu2019review}. Hence, the spatial distribution of land cover has a strong impact on the surface temperature~\citep{chen2017impacts}. }\highlight{Surface material also impact the water runoff and increase the risk of flooding. Sidewalks and roads form the main part of the urban ground surfaces. Today, the majority of the sidewalks are covered with impermeable materials which prohibit the infiltration of the water into the underlying soil, increase both the magnitude and frequency of surface runoffs~\citep{shuster2005impacts,bell2019modeling}, reduce the groundwater recharge, and negatively impact the water quality. The excessive use of impervious surfaces is shown to be the primary cause of the Combined Sewer Overflows (CSOs), which can lead to massive pollution of natural bodies of water and street flooding~\citep{joshi2021not}.}
\highlight{Aside from the mentioned impacts, sidewalk pavements can also lead to public health hazards such as outdoor falls, or pose a barrier to walkability and accessibility of public spaces, specifically for the more vulnerable population and wheelchair users~\citep{talbot2005falls, CLIFTON200795,AGHAABBASI2018475}. Studies show that uneven surfaces, indistinguishable surface colors, and low-friction materials contribute to the high incidence of outdoor falls in elderly populations~\citep{chippendale2015, thomas2020keep}.}

Despite the substantial economic, environmental, public health, and safety implications of sidewalk pavements~\citep{muench2010greenroads,van2015towards, estoque2017effects}, most cities, even in industrialized economies, still lack information about the location, condition, and paving materials of their sidewalks~\citep{deitz2021squeaky}. The lack of data creates barriers to understanding the real extent of the environmental and social impacts of using different materials and inhibits our ability to take a complex system approach to sustainability assessment~\citep{van2015towards}. For instance, studies show a significant intra-urban variability of the urban thermal environment due to the street-level heterogeneity of paving materials~\citep{agathangelidis2020urban}. However, the data scarcity makes it challenging to measure this variability across different neighborhoods~\removed{the impact of different pavement materials on
surface UHI} and consequently, \removed{develop}~\highlight{impedes the development of} a sustainable and resilient mitigation response plan~\citep{ akbari2008urban, li2013relationship,yang2019local}. In the absence of fine-scale data, studies mainly rely on remote sensing images; however, the high-resolution aerial images are both spatially and temporally sparse~\citep{zhang2009bi}, requiring researchers to use a variety of data aggregation and extrapolation techniques to fill in the missing data, which can~\highlight{lead to high bias and} hurt the validity of the final results\removed{and lead to high bias}. 

Collecting comprehensive and fine-scale sidewalk data using conventional methods is time-consuming and cost-prohibitive. Recent technological innovations in data collection opened new frontiers for research on public space and pedestrian facilities, creating opportunities to track features of interest at higher temporal frequencies and more granular geographic scales~\citep{glaeser2018big}. 
The use of street-level images in urban analysis has gained popularity since the introduction of Google~Street~View~(GSV)~\citep{anguelov2010google} and Microsoft Street Slide~\citep{kopf2010street}, services that provide panoramic images captured by cameras mounted on a fleet of cars. Concurrently, developments in machine learning and computer vision applied to these new datasets have enabled novel research directions to measure the ``unmeasurable'' in urban built environments~\citep{ewing2009measuring}, including sidewalks~\citep{frackelton2013measuring, ai2016automated}.

In this work, 
we address this data gap and take a step towards exploring the surface of our cities through CitySurfaces, a framework aimed at generating city-wide pavement material information by leveraging a collection of urban datasets. 
We combine active learning and computer vision-based segmentation model to locate, delineate, and classify sidewalk paving materials from street-level images.
Our framework adopts a recent high-performing segmentation model~\citep{tao2020hierarchical}, which uses hierarchical multi-scale attention combined with object-contextual representations.
To tackle the challenges of high annotation costs associated with dense semantic label annotation, we make use of an iterative, multi-stage active learning approach, together with a previously acquired sidewalk inventory from Boston, which lists the dominant paving material for a given street segment.
We demonstrate how the trained segmentation model can be extended with additional classes of materials with noticeably less effort, making it a versatile approach that can be used in cities with varying urban fabrics and paving materials.
To show the generalization capabilities of CitySurfaces, we employ our framework in the segmentation of street-level images from four different cities: Brooklyn, Chicago, Washington DC, and Philadelphia, none of which were included in the training process. Figure~\ref{fig:overview} highlights how different pavement materials are spatially distributed in three cities.

Our contributions can be summarized as follows:
\begin{itemize}[noitemsep]
    \item We present CitySurfaces, a deep-learning-based image segmentation framework for large-scale localization and classification of sidewalk paving materials.
    \item We adopt an active learning strategy to significantly reduce pixel-level annotation costs for training data generation, and yield increased segmentation accuracy.
    \item We conduct extensive experiments using street-level images from six different cities demonstrating that our model can be applied to cities with distinct urban fabrics, even outside of the domain of the training data. 
    \item \removed{We make the results of our material classification in the six selected cities publicly available} \highlight{We make publicly available our model as well as the results of our material classification in the six selected cities~\footnote{\url{https://github.com/VIDA-NYU/city-surfaces}}}.
\end{itemize}

This paper is organized as follows: Section~\ref{data} describes the main data sources of our framework; Section~\ref{citysurfaces} describes the CitySurfaces framework; Section~\ref{results} summarizes our results; Section~\ref{discussion} highlights challenges and limitations; and Section~\ref{conclusion} presents our conclusion.

\section{Data description} 
\label{data}

Aware of the fact that manually labeling images is a time-consuming task, our proposed framework leverages a unique dataset that describes the material of sidewalks in Boston.
We combine that data with the street-level images to create the training data for our semantic segmentation model. 
%
Next, we describe both data sources.

\subsection{Boston sidewalk inventory}
\label{data:boston} 

\begin{figure}[t]
\centering
\includegraphics[width=0.5\textwidth]{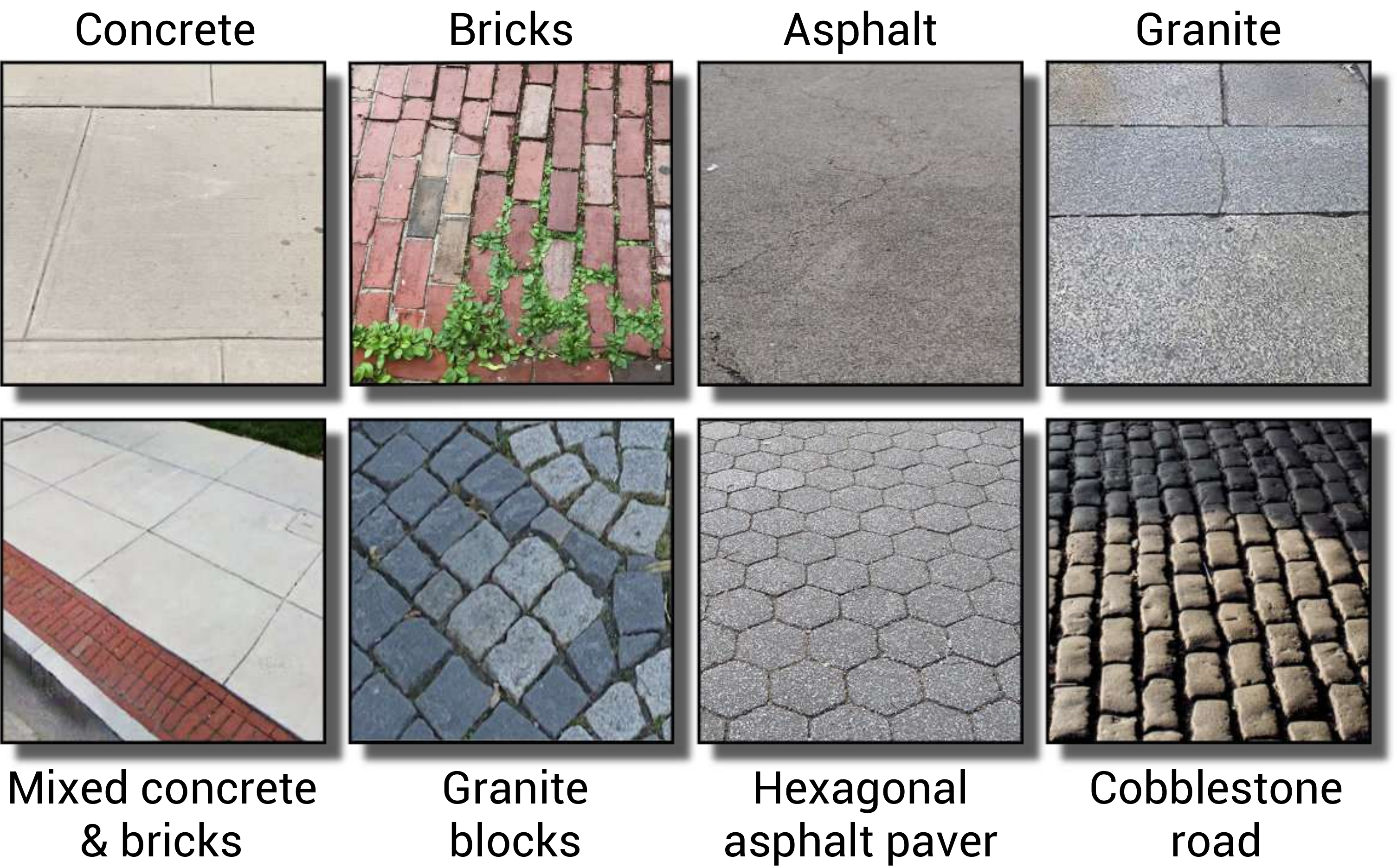}
\caption{\removed{The eight classes of surface materials used in our framework. The top row presents standard and prevalent materials found in most of the cities. The
mixed class is the next popular one and the three remaining materials in the second row have distinct use. Cobblestone is street pavement material, mostly
seen in historic neighborhoods.} \highlight{The eight classes of surface materials used in our study. Top: standard and prevalent materials, Bottom: materials with distinct use.}}
\label{fig:materials}
\end{figure}

The sidewalk inventory~\citep{Bostonswinv} is part of the Boston~Pedestrian~Transportation~Plan~\citep{bostonregional} and describes sidewalk features, including geographic coordinates and paving materials collected via manual field visits. The material attribute describes the dominant surface material of each street segment (either concrete, brick, granite, a mix of concrete and brick, or asphalt). Figure~\ref{fig:materials} illustrates patches of these five materials; the other three extra materials (granite block, cobblestone, hexagonal pavers) shown in the image were not recorded in the Boston dataset but were later manually added to our classes, as we will discuss in Section~\ref{block(c)}. We grouped the street segments by materials, using the geographic coordinates of the paving materials in the Boston inventory,  and used it to assign an overall image class to the street-level images to guide the annotation process.


\begin{figure}[t]
\centering
\includegraphics[width=0.5\textwidth]{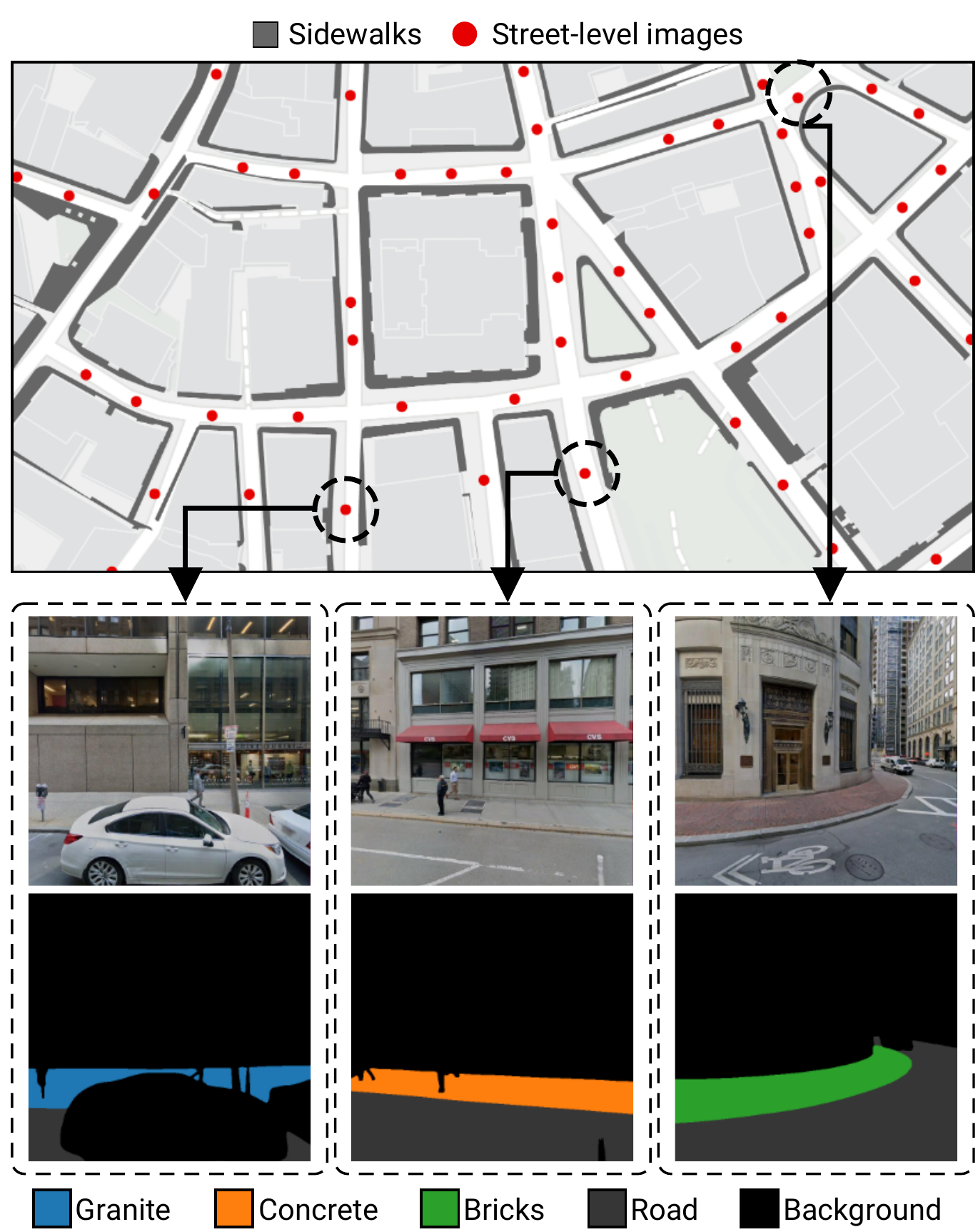}
\caption{Examples of sampled points in Boston to obtain street-level images. Three different sampling locations are highlighted and for each location, the street-level image as well as the prediction result of the model is depicted.}
\label{fig:gsv-diverse}
\end{figure}

\subsection{Street-level images}
\label{data:gsv}

Street-level image usage in urban analysis has gained popularity with the introduction of Google Street View (GSV)~\citep{anguelov2010google} and Microsoft Street Slide~\citep{kopf2010street}, services that provide panoramic images captured by specifically designed cameras mounted on a fleet of vehicles. These new data sources enable new questions and study designs for urban planning and design, urban sociology, and public health~\citep{griew2013developing,yin2015big,mooney2016use}.  The GSV API can retrieve street-level images via geographic coordinates and allows users to adjust camera settings such as the heading, field of view (FoV), and pitch.

\begin{figure*}[t!]
\centering
\includegraphics[width=0.9\linewidth]{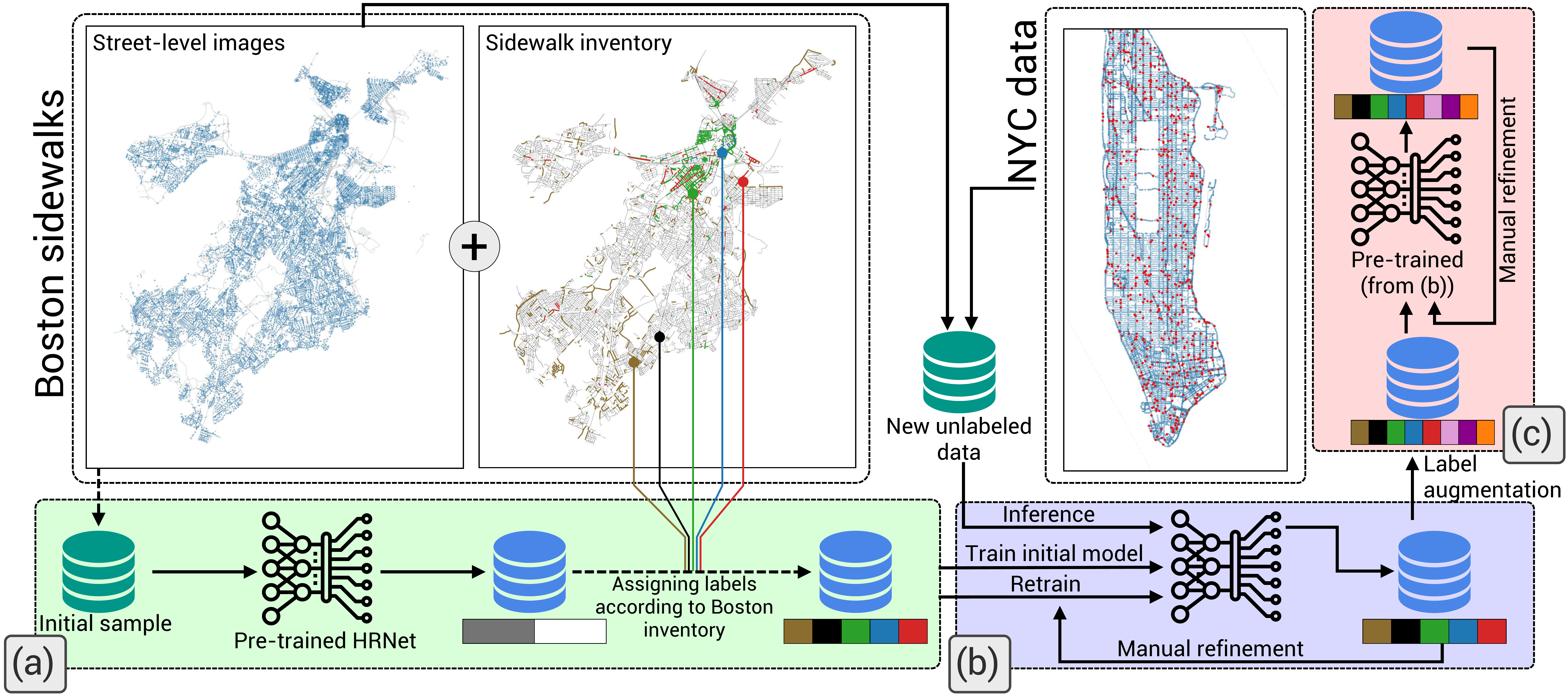}
\caption{\removed{CitySurfaces workflow: \textbf{Block (a):} Creating the initial ground truth labels using the Boston sidewalk inventory and GSV images. A sample of images is chosen and fed to a pre-trained HRNet, which outputs annotation labels containing two classes of interest: roads and sidewalks. The labels are then manually refined so each paving sidewalk material is represented using a distinct color which will form our ground truth set; \textbf{Block (b)} We iteratively choose representative and informative samples to retrain the segmentation model, which can classify five classes of surface materials, plus roads. The model is run on new data, and the segmentation accuracy is used as criteria to guide choosing new samples from Boston and later Manhattan images. In each stage, the most representative and informative test data are chosen, and the annotations are manually refined and added to the training set to retrain the network; \textbf{Block (c)} Introducing three new classes of materials that initially were not part of the Boston inventory, but were among the standard NYC sidewalk materials. The pre-trained model from block (b) is retrained on the newly annotated image data, including the three new classes. The final model can classify eight classes of different materials, including cobblestone streets.}~\highlight{CitySurfaces workflow. Block (a): Creating the initial ground truth labels using the Boston sidewalk inventory and GSV images. A sample of unlabeled images is fed to a pre-trained HRNet, which outputs annotation labels containing two classes of interest: roads and sidewalks. The labels are manually refined to represent the five sidewalk paving classes, forming our ground truth set; Block (b): Training the base model to classify five classes of surface materials, plus roads. The data from block (a) is used for the first stage of training. The model is then iteratively retrained for multiple stages on new samples. In each stage, the most representative and informative samples are chosen, and the annotations are manually refined and added to the training set to retrain the network; Block (c): Introducing three new classes of materials. The pre-trained model from block (b) is retrained on the newly annotated image with three new classes. The final model can classify eight classes of different materials.}}
\label{fig:workflow}
\end{figure*}

We use the OSMnx library~\citep{boeing2017osmnx} to obtain the Boston street network and query the GSV API for street-level images at a fixed interval~(5 meters), excluding major highways and tunnels. We acquire the compass bearing of each street to set the camera heading to be perpendicular to the street, thus looking directly at left and right sidewalks. The pitch was set to $0^{\circ}$ with an FoV of $80^{\circ}$. To create a more diverse training set, for 35\% of the training data, we use different combinations of headings (pitch $\in[-10^{\circ}, -20^{\circ}$], and FoV$\in [60^{\circ}, 70^{\circ}$]), to have sidewalk images taken at varying angles and perspectives. Figure~\ref{fig:gsv-diverse} illustrates sampled street segments in Boston, together with their image-level annotations. In order to train our framework, 3,500 Boston images were obtained, and later 2,000 images from New York City (NYC) were added to the pool of initially unannotated data. We excluded images with no sidewalks as well as those where more than 80\% of the sidewalks were occluded. The final set had a total of 4,300 images.


\section{CitySurfaces}
\label{citysurfaces}

CitySurfaces adopts an active learning approach for the semantic segmentation of sidewalk paving materials. Using this framework, we aim to: 1) Train a model that can classify five different paving materials plus asphalt roads; 2) Extract information about sidewalk materials of a city for which no ground truth sidewalk inventory exists (e.g., NYC); and 3) Extend the model to classify additional classes of materials so that it can be applied to a more general set of cities.

Active learning aims at achieving high accuracy while minimizing the amount of required labeled data. The main hypothesis is, if we allow the model to choose the training data, it will perform better with fewer labeled instances~\citep{settles2009active}. Through iteratively selecting the most informative or representative images to be labeled, fewer labeled instances are required to achieve similar performance compared to randomly selecting a large sample as training data and annotating all of it at once \citep{bloodgood2014method, huang2010active}. 

In general, our multi-stage workflow is different from previous works in active learning for semantic segmentation in two important ways: First, our sample selection method is not fully automated; we use the uncertainty measure to filter the pool of unlabeled data in each stage, but we also use domain expertise for selecting a sample of images to be annotated and added to the training set in the next stage.
Second, our query frequency is ten epochs (each epoch is a pass through all training data). The conventional approach in active learning is to select new samples (query) every iteration, which can work in cases where the cost of annotation is not high or in experimental studies that work with already annotated images to advance the field and develop new query algorithms, as is the case with most of the already published works in active learning for semantic segmentation, where they use datasets such as Cityscapes~\citep{cordts2016cityscapes} or ADE20k~\citep{zhou2017scene}. However, since no annotated dataset exists for sidewalk materials, we have to annotate every new sample we choose during the training process, and it is impractical to annotate a new sample for every iteration~\citep{kim2020message}. To overcome this, we adopt a multi-stage framework and annotate a new sample at the end of each stage, where each stage consists of ten epochs.

Our workflow has three major blocks as illustrated in Figure~\ref{fig:workflow}: Block (a) creating initial training labels; Block (b) training a material segmentation model and; Block (c) extending the model to segment three additional classes from NYC standard materials. In this section, we first describe the different blocks of the workflow in detail, followed by a description of the semantic segmentation model.
The training process and experiments were executed on 4 NVIDIA P100 GPUs with 12 GB of RAM each.

\subsection{Block (a): Initial image annotation}

To start the training process, we need a set of annotated images. To obtain the annotated data, we randomly sample 1,000 images from a pool of unlabeled Boston street-level images and feed that sample into HRNet-W48~\citep{sun2019high, wang2020deep} model pre-trained on Cityscapes~\citep{cordts2016cityscapes} and get the initial segmentation results (Figure~\ref{fig:workflow}(a)). The model outputs 19 classes from which we only keep roads and sidewalks. 
To generate an initial set of labeled data, we make use of the Boston Sidewalk Inventory (detailed in Section~\ref{data:boston}). We first query for the street segments of the images in our initial sample and modify the label to match the audited pavement from the inventory. Effectively, we are ensuring that, instead of having a general \emph{sidewalk} class outputted by the pre-trained HRNet, our image set will have annotations according to the ground truth inventory data (e.g., concrete, bricks).
We then manually refine them to account for the pre-trained model's prediction errors. In the initial training set, we restrict our sampling to images where the sidewalks mainly consist of a single material and eventually move to more complex material configurations in later stages. The final annotated images were split into 80\% training and 20\% validation to train the model in block (b). 

\subsection{Block (b): Model training on Boston and NYC}
\label{blockb}
In the second block of the framework (Figure~\ref{fig:workflow}(b)), we train an attention-based model (detailed in Section~\ref{model}) using the labeled images from block (a). Our training step initially uses 800 images for training, and 200 images for validation, with a batch size of 8, SGD for the optimizer, momentum 0.9, weight decay $5e^{-4}$, and an initial learning rate of 0.002. 
We train the model in a multi-stage framework, where each stage consists of ten epochs. In each stage, we choose the epoch with the highest mIoU on the validation set.
At the end of each stage, we make two decisions: 1) we select the best model considering all epochs of the current stage; and 2) we analyze the quantitative and qualitative results of the model to guide sampling the \emph{new} \highlight{addition to the} training data. In particular, we analyze the confusion matrix, similarity matrix, as well as the top 10\% of predictions with the highest mIoU and the top 20\% of failures, obtained from the validation phase of the best epoch.
The weights of the best model in the current stage are then used to initialize the model in the next stage with more training data. This restating scheme of SGD with the best solution of the previous stage is useful in increasing the chances of finding better solutions in the current stage.

To sample new images, we employ two strategies: i) Uncertainty in predicting unlabeled images: We make use of the model's uncertainty estimations on unlabeled data and select the images that were most challenging for the model to predict; and ii) Performance on validation set: By examining per-image IoU, uncertainty, and error rates of the images from failure and success cases together with confusion matrices, we construct a set of sample images to be used as inputs for finding similar unlabeled images. A more detailed explanation of these two techniques is provided in~\ref{appendixa}. 

\begin{figure}[b!]
\centering
\includegraphics[width=0.5\textwidth]{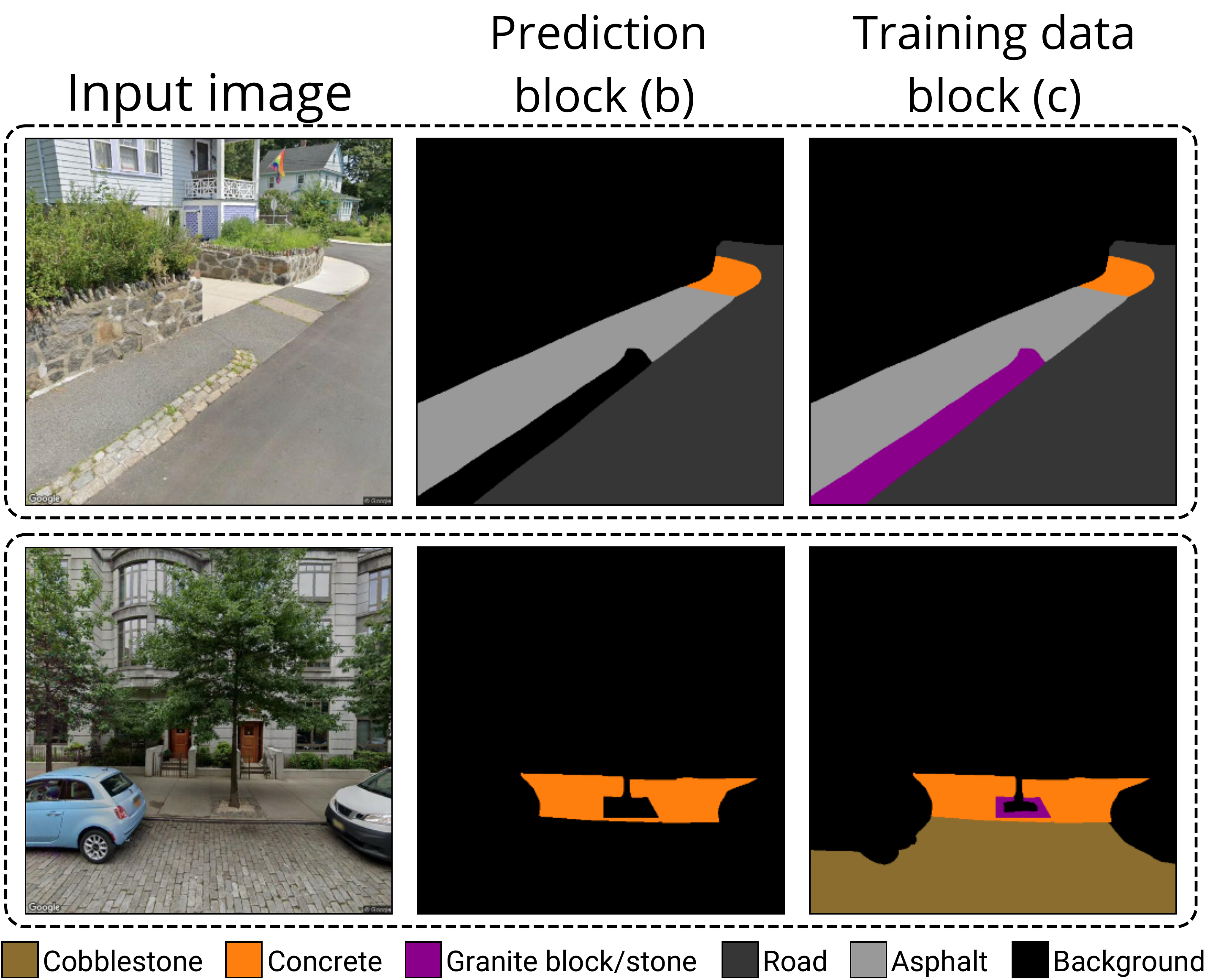}
\caption{Examples of how the annotation labels with additional classes were created from the output of the model in block (b) of our framework. The model trained in block (b) classified granite blocks and cobblestone as background, leaving smooth and clear boundaries, which helps to augment the labels with new classes during manual refinement and train a model that can classify eight different materials (block (c) of the framework).}
\label{fig:newmaterials}
\end{figure}

Following the sample selection strategies, we retrieve 300 unlabeled images, apply the current model on these new unlabeled images to generate a prediction, and then modify the predicted labels to add them to the overall training set, such that the segmentation model is trained on more samples of hard-to-segment images. To improve model generalization, in the third stage, we begin including images from Manhattan, which has a different urban fabric and more diverse forms and types of paving materials, in the pool of unlabeled data. Since no ground truth data exists for Manhattan, to create the ground truth label, we need to have a model with reliable performance to create the base annotation. We chose the third stage since the model reached a reliable performance (83\% mIoU) in detecting the main classes, and outputs had clear borders compared to the other two stages. The selected images from Manhattan were fed to the model, and the results were corrected and refined using feedback from the domain expert and added to the training dataset. 
The segmentation model is then trained on the combined set of the initial and newly annotated data (1,100 images), initialized with the weight from the best epoch of the previous stage. This procedure is iterated for five stages (at which point we observe no further notable improvements). The model at the final stage was trained on 2,500 images (Figure~\ref{fig:workflow}(b)), and achieved 88.6 \% mIoU on the held-out test set. 

\begin{figure*}[t!]
\centering
\includegraphics[width=1.0\linewidth]{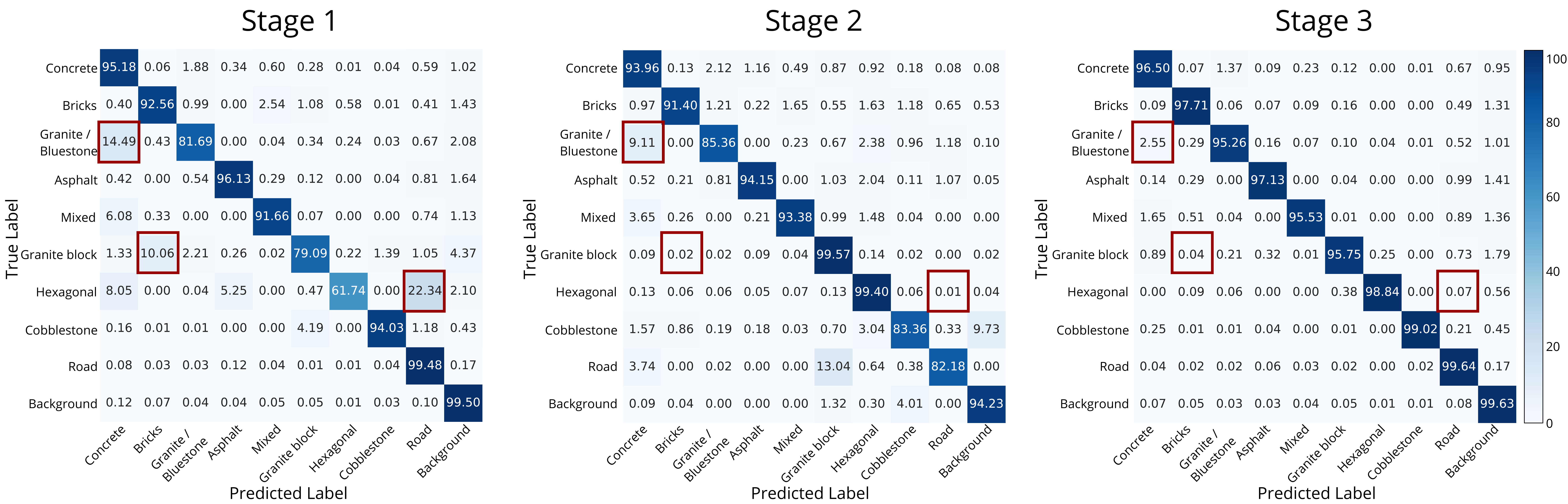}
\caption{Confusion matrices for the three stages of the extended model. These results guided sample selection and signaled which type of images should be included in the training data for the next stage. Notice the improvement of the predictions for hexagonal pavers, granite block, and granite/bluestone (highlighted in red).
}
\label{fig:matrix}
\end{figure*}

\subsection{Block (c): Including additional materials from NYC}
\label{block(c)}

Once the model in block (b) attains sufficiently accurate segmentation performance, we extend it by adding three additional classes (Figure~\ref{fig:workflow}(c)). The three additional classes are granite blocks, hexagonal pavers, and cobblestone. These materials are standard sidewalk materials in the NYC street design manual~\citep{new2020street}. While granite blocks and cobblestones were also observed in Boston, they were not included in the Boston sidewalk inventory. Since the original model in block (b) was not trained to detect these materials, they are initially either classified as background (mostly granite blocks and cobblestones) or misclassified (mostly hexagonal pavers) as other visually similar materials. To collect street-view images that have these new materials, we follow the NYC and Boston street design manuals \citep{bostoncomplete,new2020street} to filter unlabeled data from the locations in which these materials can be found. For example, hexagonal pavers (NYC only) are typically used on sidewalks adjacent to parks and open spaces, and cobblestones are used in historic districts. 

We select a total of 800 images that contain these new classes to be iteratively sampled for training, 150 additional images for the validation set, and 200 images for the held-out test set. Annotating the new image set consumed fewer resources as compared to the initial annotations since smooth model predictions typically leave clear boundaries, which only needed to be assigned the appropriate label (see Figure~\ref{fig:newmaterials}). The newly generated set of labels was used to train the model by initializing the architecture with model weights in block (b) and only replacing the final softmax layer instead, to produce ten output channels (corresponding to eight paving materials, plus the road, and background). At the end of each stage, we select a new sample of unlabeled images following the same process explained in Section~\ref{blockb}, run them through the model, obtain segmentation predictions, refine the results, and retrain the model. In total, 726 additional images were added to the training set, and in the final stage, the model was trained on 3,226 images (2,500 from block (b) + 726). We halt the training in stage 3 after 30 epochs, and test the model on the held-out test set~(Figure~\ref{fig:workflow}(c)). Figure~\ref{fig:matrix} shows the confusion matrices for all three stages of our extended model, illustrating model performance as a function of the amount of training data. These matrices were also used in part to guide the sampling of images to annotate.

\begin{figure}[t!]
\centering
\includegraphics[width=0.5\textwidth]{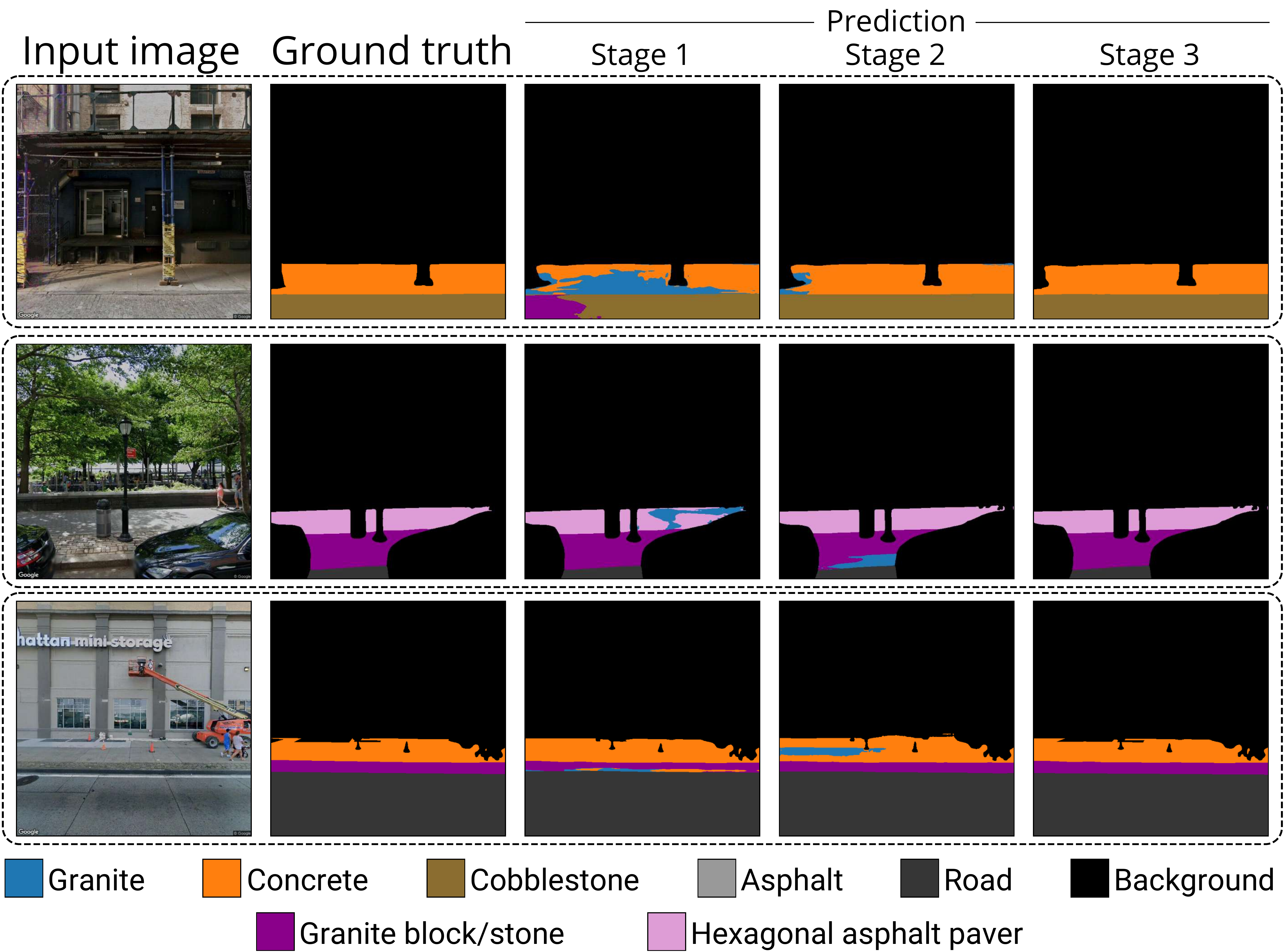}
\caption{Evolution of the block (c) extended model's inference through different training stages.}
\label{fig:cycles}
\end{figure}

Using the described method, model performance increases from 74.3\% mIoU to 88.6\% for the base model (block (b)) and to 90.5\% in the extended model (block (c)), with the manual refinement time decreasing from 25 to 4 minutes per image. Figure~\ref{fig:cycles} depicts the evolution of the segmentation results of block (c) through the active learning stages. The model outputs more refined boundaries and significantly less noise in later stages; thus, significantly less time is needed to modify the newly annotated data as the stages go on. In each stage, the model is initialized with the weights from the previous stage.

\subsection{Semantic segmentation model} 
\label{model}

\begin{figure}[b!]
\centering
\includegraphics[width=0.5\textwidth]{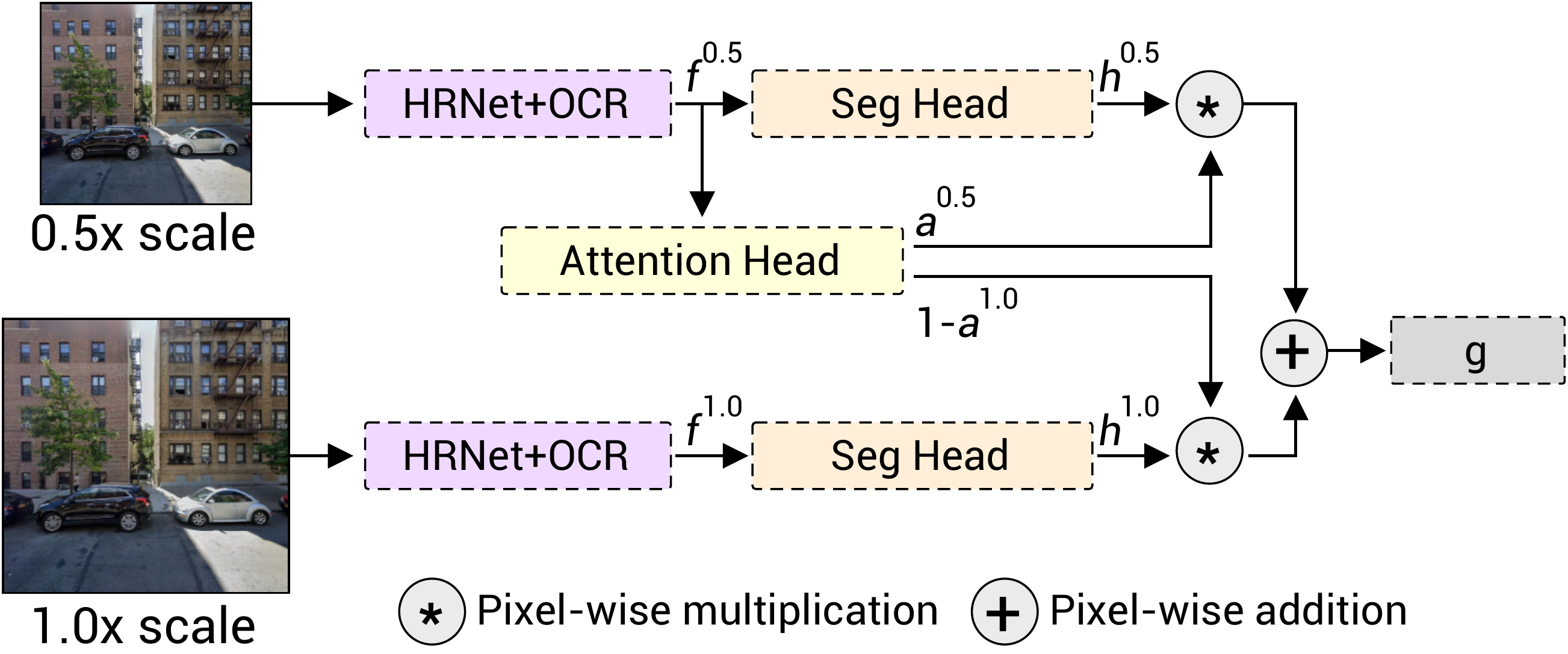}
\caption{The general architecture of the hierarchical multi-scale attention (HMSA) based semantic segmentation method \citep{tao2020hierarchical}. The inputs are images from two scales. The network learns the relative attention between scales and hierarchically applies the learned attention to combine the results from two segmentation heads and make a prediction.}
\label{fig:attention}
\end{figure}

For the semantic segmentation task (blocks (b) and (c)), we adopt the Hierarchical Multi-Scale Attention~\citep{tao2020hierarchical} and fine-tune the parameters on our dataset. To train the model, following~\citet{zhu2019improving}, we employ class uniform sampling in the data loader, which chooses equal samples for each class for handling the class imbalance, since some classes like road and background are almost present in all images, whereas classes like cobblestone and hexagonal pavers are not that prevalent. 
The Region Mutual Information (RMI) loss \citep{zhao2019region} was employed as the primary loss function.  RMI takes the relationship between pixels into account and uses the neighboring pixels around each pixel to represent it instead of only relying on single pixels to calculate the loss. We run different experiments with and without the RMI loss function for the main segmentation head. In the absence of RMI, standard cross-entropy loss was used instead. The model under the same setting, but without RMI loss, performed slightly worse (89.84) compared to the one where RMI loss was used (90.51). Figure~\ref{fig:attention} presents an overview of the architecture. Next, we describe the network's architecture in more detail.

\subsubsection{Backbone}
\label{hrnet}
We chose HRNet-OCR~\citep{yuan2019object} as the backbone. The network comprises HRNet-W48~\citep{sun2019high,wang2020deep} and adds Object-Contextual Representations \citep{yuan2019object} to further augment the representation extracted by the HRNet. The final representation from HRNet-W48 works as the input to the OCR module, which computes the weighted aggregation of all the object region representations to augment the representation of each pixel. The weights are calculated based on the relations between pixels and object regions. The augmented representations are the input for the attention model described next. 

\subsubsection{Attention model}

The model is mainly based on Share-Net~\citep{chen2016attention}. Suppose an input image is resized to several scales, i.e., $s \in \{ 1,...,S\}$. Each scale is passed through the backbone part (HRNet-W48+OCR), and we can get the output feature $f_{i,c}^s$. For the feature, $c \in \{1, ..., C\}$ ($C$ is the number of classes of interest, and $i$ ranges over all the spatial positions). As shown in Figure~\ref{fig:attention}, the features then go through two heads, one for attention generation and the other for segmentation. The features $f_{i,c}^s$ are resized for different scales to have the same resolution (with respect to the finest scale) using bilinear interpolation before passing the model heads. For the attention head, we generate the learned weights for $f_{i,c}^s$ which is represented by $a_{i,c}^s$. This weight is integrated into the initial output $h_{i,c}^s$ from the segmentation head, and we have:
\begin{equation}
    g_{i,c}^s = a_{i,c}^s * h_{i,c}^s
\end{equation}
in which $g_{i,c}^s$ is the final output score map for scale $s$, and $*$ here represents the pixel-wise multiplication.

In the model, the combination of score maps is similar to \citep{tao2020hierarchical} to make the flexible scales during inference time possible and improve the training efficiency. During the training, we only need to train with two adjacent scales (as shown in Figure~\ref{fig:attention}). During testing, weights for the network are shared for each adjacent scale pair.

To be more specific, suppose the two selected adjacent scales are $1x$ and $0.5x$ (the final selected scales during training in the model are $0.5x$, $1x$, and $2x$) to obtain the pair of scaled images for the model input. For inference, we can hierarchically and repeatedly use the learned attention to combine $N$ scales of predictions together. Precedence is given to lower scales since they have a more global context and can choose where predictions need to be refined by higher scale predictions. The final combination principle for these adjacent scales is defined as:
\begin{equation}
    g_{i,c} = a_{i,c}^0.5 * h_{i,c}^0.5 + (1-a_{i,c}^0.5)* h_{i,c}^1
\end{equation}

The hierarchical mechanism used in the model coupled with the powerful HRNet-OCR backbone provides a robust architecture for the challenging task of material classification \emph{in the wild}. 
%


\begin{figure*}[t!]
\centering
\includegraphics[width=0.95\linewidth]{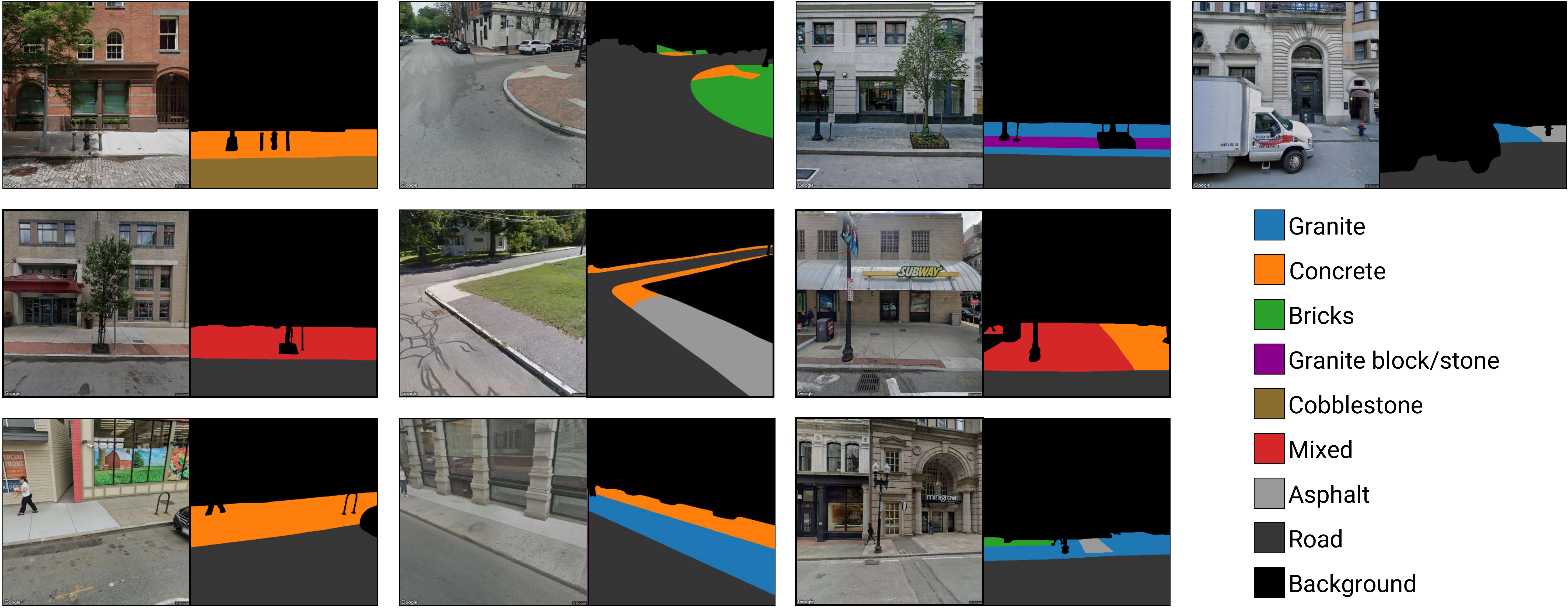}
\caption{Predictions of the model on the held-out test set. Fine details and boundaries of objects like poles, plants, wooden sticks, and fire hydrants are very precisely predicted. The model also segmented curb cuts (line 1 - column 2), different instances of the same material (3-1), (3-3), and visually similar materials of different classes (1-4). 
}
\label{fig:results}
\end{figure*}

\section{Results}
\label{results}
In this section, we present the results of applying our trained model on the held-out test set. We do not rely on pixel-level accuracy in evaluating the model since sidewalks comprise a relatively small portion of each image, while road and background can occupy more than 70\% of most images, resulting in a significant class imbalance. This class imbalance creates an arbitrary high pixel-level accuracy, which is not a fair representation of the model's performance.

\subsection{General evaluation metrics}
Table~\ref{tab:Results} presents class-level evaluation metrics, the mean Jaccard index (IoU), precision, and recall for the final model. The model outputs ten classes in total, seven classes of sidewalk pavings, one extra class of street pavings - cobblestone - plus road and background. Excluding road and background, the model achieved 88.37\% accuracy, with hexagonal asphalt pavers and asphalt sidewalks having the highest accuracy measures. Overall, half of the pavement classes have IoU above 90\%. Concrete, the most prevalent and versatile material, can be classified with 88.7 accuracy. A robust result considering the high within-class variation (i.e., it comes in various colors and textures). 
Granite/bluestone and granite block have the lowest accuracy (81.09 and 82.92 respectively). This can be partially explained by their visual similarity to dark concrete (or wet concrete), potentially leading to more false positive predictions.

\removed{Figure~\ref{fig:results} highlights some important cases, especially considering the \emph{in the wild} nature of GSV images. We can notice that CitySurfaces performs well in detecting boundaries between fine objects, like poles and plants, even in shadowed scenes (line 1 - column 1, 1-3, 2-1). }\highlight{Figure~\ref{fig:results} illustrates some examples of the model's prediction, highlighting its performance in detecting boundaries between fine objects, like poles and plants, even in shadowed scenes (line 1 - column 1, 1-3, 2-1).}
The model can also detect curb ramps in most scenes, even though it was not specifically trained with such a goal (1-1 and 2-2).
\removed{Figure~\ref{fig:results} (1-2) shows an area with patches of different materials and how the model can accurately classify each type. }\highlight{Figure~\ref{fig:results} (1-2) shows an example in which the model accurately classified a sidewalk segment with patches of different materials.}
We can also see the model performance in distinguishing between visually similar materials (1-4, 3-2), \removed{as well as visually different types of the same material (3-1). }\highlight{as well as different variation of the same material such as (3-1) where two visually distinct concrete slabs are classified correctly.}

\begin{table}[t!]
\centering
\caption{Evaluation metrics on the held-out test set.}
\label{tab:Results}
\begin{tabular}{c|c|c|c}
\hline
Label & IoU & Precision & Recall \\ \hline
Concrete & 88.69 & 0.95 & 0.93 \\
Brick & 91.79 & 0.95 & 0.96 \\
Granite/Bluestone & 81.09 & 0.85 & 0.95 \\
Asphalt & 92.58 & 0.96 & 0.97 \\
Mixed & 86.11 & 0.93 & 0.93 \\
Granite Block/Stone & 82.92 & 0.94 & 0.88 \\
Hexagonal Asphalt Paver & 92.81 & 0.98 & 0.95 \\
Cobblestone & 90.95 & 0.94 & 0.96 \\
Road & 99.01 & 0.99 & 1  \\
Background & 99.16 & 1 & 1 \\ \hline
\textbf{mIoU} & \multicolumn{3}{c}{90.51} \\ \hline
\textbf{mIoU (eight main classes)} & \multicolumn{3}{c}{88.37} \\ \hline
\end{tabular}
\end{table}

\subsection{Evaluating the generalization capabilities of CitySurfaces}

To demonstrate the generalization capabilities of CitySurfaces,\removed{we selected Chicago, Washington DC, Philadelphia, and Brooklyn (NYC's borough), to test the performance of the proposed approach on each city.As previously mentioned, we highlight that CitySurfaces was only trained using images from Boston and Manhattan. }\highlight{ we tested the performance of our approach on samples from Chicago, Washington DC, Philadelphia, and Brooklyn (NYC borough), which were not part of the training data.}
\begin{table}[t!]
\centering
\caption{Evaluation metrics on samples from the selected cities (outside of training domain).}
\label{tab:cityresults}
\begin{tabular}{l|l|c}
\hline
City & \multicolumn{1}{c|}{mIoU} & Mean Per-Segment Accuracy \\ \hline
Brooklyn & 86.12 & 87.09 \\
Chicago & 84.31 & 86.52 \\
Washington DC & 82.61 & 84.27 \\
Philadelphia & 82.81 & 83.46 \\
\hline
\end{tabular}
\end{table}
We randomly sampled 200 street segments from each city, and obtained their corresponding street-view images, at every five meters of each segment, from the left and right sides of the sidewalks. After data cleaning and pre-processing, we were left with roughly 600 images per city; these images were annotated using the model in block (b), then \emph{manually} checked and refined to create the test set. Table~\ref{tab:cityresults} shows the results of applying CitySurfaces on these test sets. \removed{both using a mIoU and a mean per-segment accuracy. This last measure was defined as a simple and more practical metric that measures whether the model could correctly detect the main dominant material in each street segment. }\highlight{We report mIoU and mean per-segment accuracy. Mean per-segment is a simple and practical metric that measures whether the model correctly detected the dominant materials in each street segment and report the average accuracy over all images in the test set.} 
All tested cities had an accuracy greater than 82\%. Brooklyn achieved the highest accuracy, since the borough's paving materials follow the same street design regulation as Manhattan, \highlight{which was part of the training data.}

\begin{figure}[t!]
\centering
\includegraphics[width=0.5\textwidth]{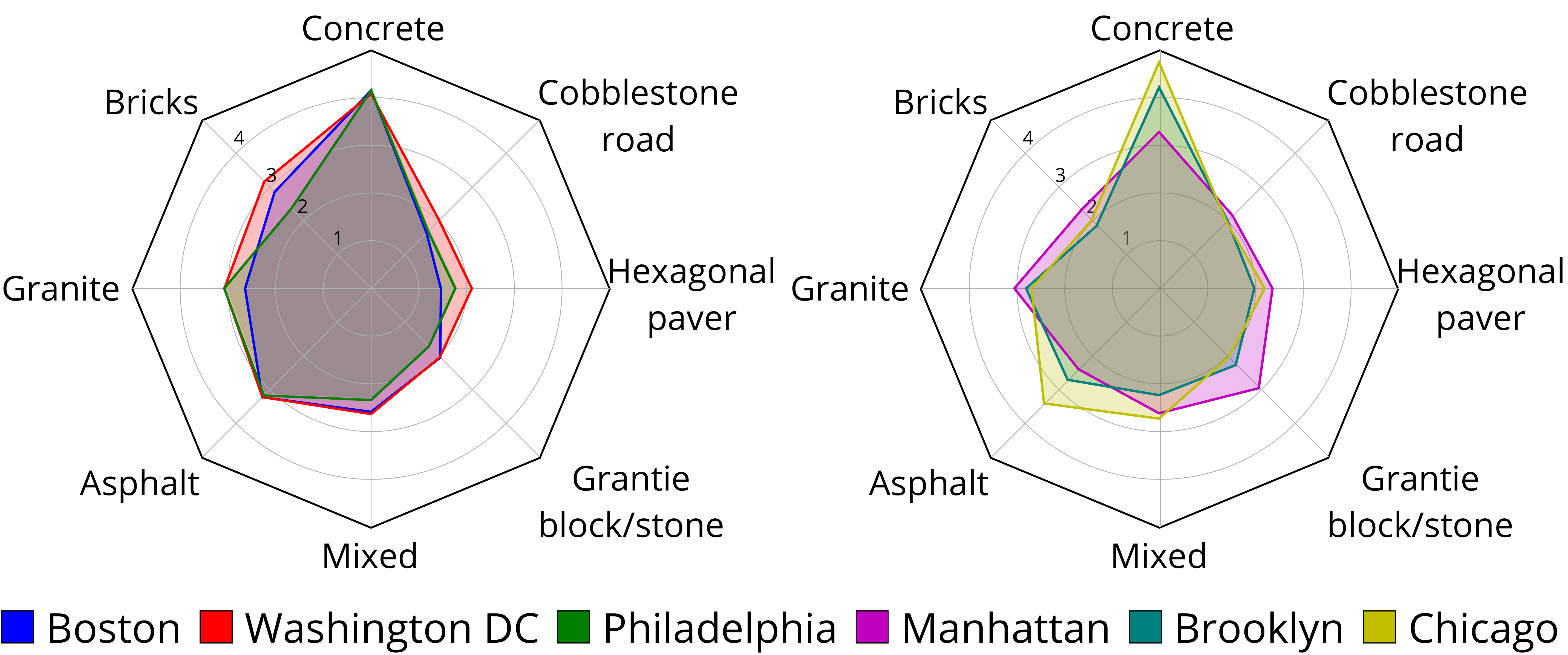}
\caption{Comparison of the distribution of detected materials in six different cities. The star plots show the log of the number of sidewalk segments identified as having a given material.}
\label{fig:radardiagram}
\end{figure}

CitySurfaces enables generating city-wide sidewalk material datasets, as illustrated in Figure~\ref{fig:results}. This allows us to compare the distribution of different paving materials in various cities.
Figure~\ref{fig:radardiagram} shows the result of this comparison. We can see that \highlight{Manhattan and Washington DC use more diverse and balanced material types.} Concrete is the dominant material in all of the cities. Chicago has the highest number of asphalt sidewalks among the selected cities; Boston, Washington DC, and Philadelphia have a similar number of asphalt sidewalks, which come second to Chicago. Asphalt sidewalks are mainly used in suburban neighborhoods; that is why dense urban areas like Manhattan and Brooklyn have the lowest number of sidewalks paved with asphalt. Another interesting observation is the higher usage of granite/bluestone in Manhattan compared to Brooklyn, two boroughs of the same city. Granite is considered an expensive and decorative material, used mainly in commercial streets or historic neighborhoods, which signals Manhattan's higher land value and income level, since maintenance and installation of decorative pavings are the owner's responsibility. 
\removed{Another interesting observation is that Manhattan and Washington DC use more diverse and balanced material types.}

\begin{figure}[t!]
\centering
\includegraphics[width=0.5\textwidth]{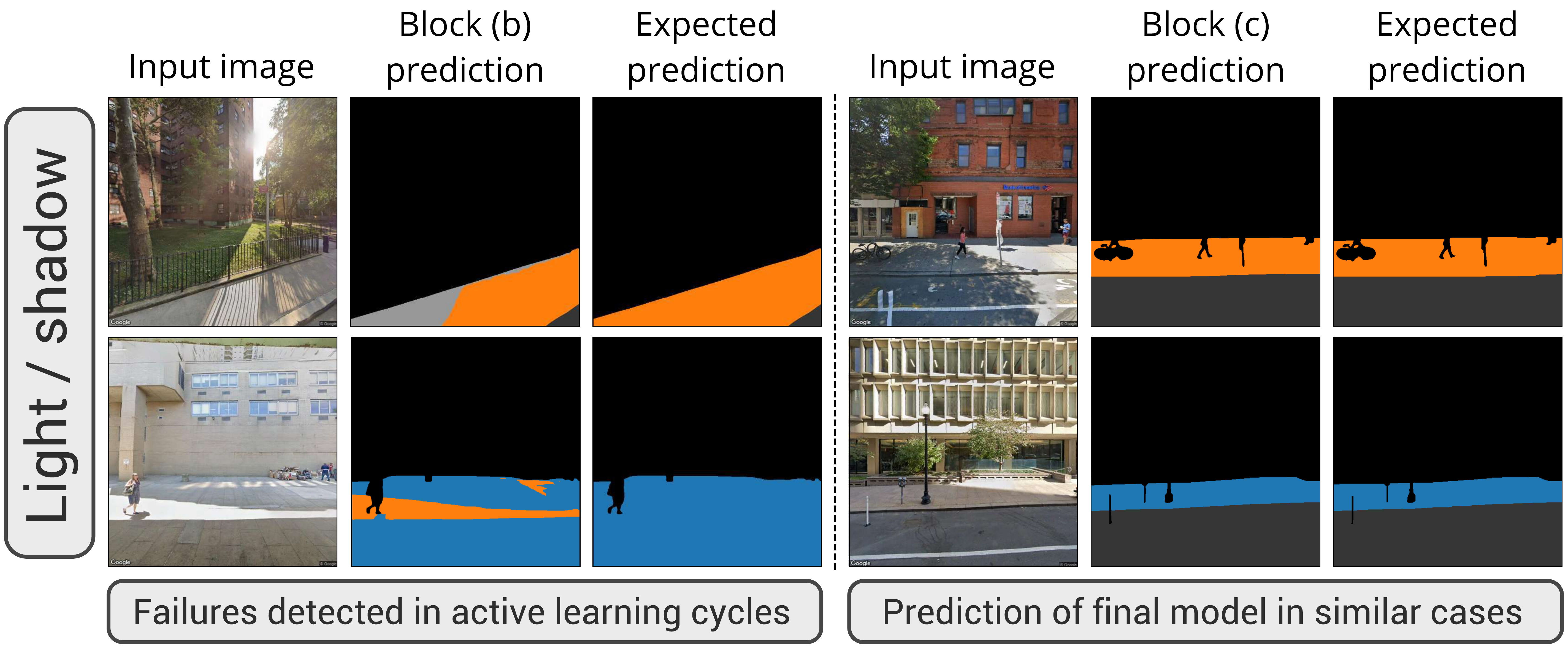}
\caption{\textbf{Left}: Exposure to direct sunlight changed the appearance of colors and texture of the paving material, \removed{making them resemble another class.} \textbf{Left top}: \removed{asphalt was predicted for the parts in shadow instead of concrete.} \highlight{Part of a concrete sidewalk under the shadow was classified as asphalt.} \textbf{Left bottom}: \removed{Concrete was predicted for the parts of the sidewalk under direct sunlight instead of granite.} \highlight{Part of a granite surface under direct sunlight was classified as concrete.} \textbf{Right}: The correct predictions of the final model in the same settings.}
\label{fig:failures_light}
\end{figure}

\section{Discussion}
\label{discussion}

\begin{figure*}[t!]
\centering
\includegraphics[width=0.93\textwidth]{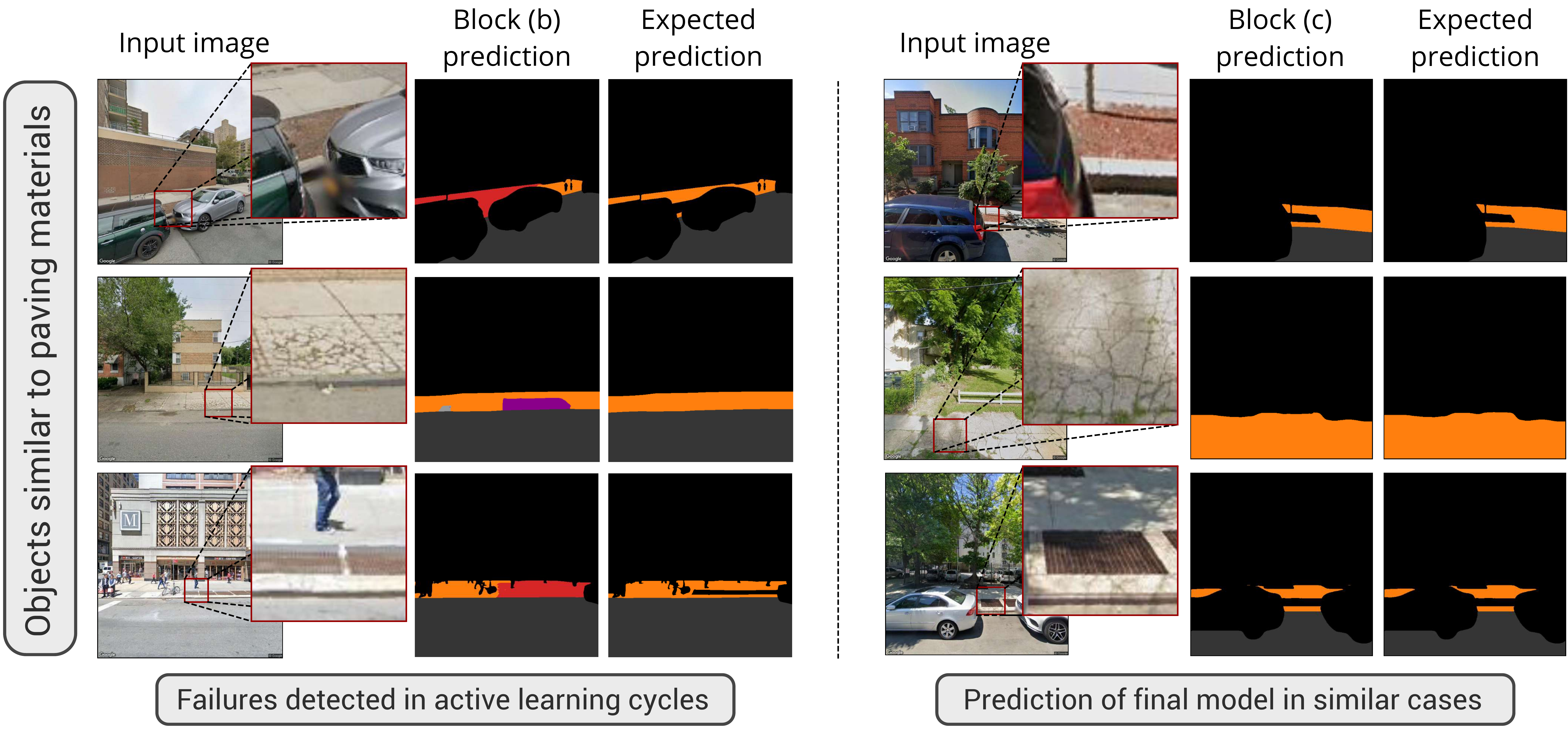}
\caption{\removed{Objects with patterns similar to different materials. Left: Random patterns created by different objects led to misclassifications. Left top: Plant pit
was classified as brick, and then model predicted mixed class instead of concrete for that part of sidewalk segment. Left middle: Broken parts of concrete
were classified as small stones. Left bottom: The brownish metal covers were classified as brick, leading to the perdition of mixed class instead of concrete.
Right: using an active learning approach, the model corrects itself and, in the final cycle, predicted correct classes in similar settings.}~\highlight{Objects with patterns similar to different materials. \textbf{Left}: Classifying failures caused by different patterns. \textbf{Left top}: Concrete alongside a furnishing zone was misclassified as mixed class since plant pit was detected as bricks, ~\textbf{Left middle}: Broken concretes were misclassified as granite blocks, \textbf{Left bottom}: Concrete was misclassified as mixed class due to the presence of brownish metal covers. \textbf{Right}: Correct prediction of the model for the similar pattern in the final cycle of active learning.}}
\label{fig:failures_material}
\end{figure*}

The specific characteristics of computing the spatial distribution of sidewalk pavement materials require experts to oversee the performance of the model and ensure that the network is correctly classifying the pavement materials. 
Through active learning process, we identified certain elements of the urban scenes that can create higher prediction confusion and lead to misclassification. 
Two main categories of patterns repeatedly observed among the failure cases were shadow/light contrasts (Figure~\ref{fig:failures_light}) and distinct objects such as metal gratings and plant pits that resemble brick from a distance (Figure~\ref{fig:failures_material}). The texture and color of different materials can appear different under shadow or extreme light, showing a higher resemblance to another material. For instance, under the shadow, concrete is classified as asphalt (Figure~\ref{fig:failures_light} - left top). 
Moreover, some patterns or objects can look similar to certain materials. For example, the model initially classified certain plant pits (Figure~\ref{fig:failures_material} - left top) or brownish metal covers (Figure~\ref{fig:failures_material} - left bottom) as bricks alongside the concrete pavement and would incorrectly predict mixed materials for that part of the sidewalk, or even small pieces of broken concrete or granite were classified as cobblestones (Figure~\ref{fig:failures_material} - left middle). Adding more images with these patterns to the training data improved the model's performance in the next stage. Some examples of the correct predictions for similar patterns are shown on the right side of Figure~\ref{fig:failures_material}.
The active learning strategy significantly helped with choosing the right data \highlight{at each stage}. Having an expert in the loop to review the results in each stage enabled identifying specific patterns that were not evident by merely analyzing the quantitative metrics of the model. 

\subsection{Challenges}
One of the key challenges of this study was handling different textures of the same object (sidewalk). Objects have defined boundaries that are easier to classify~\citep{jain2018recognizing}. However, similar textures can appear on multiple objects. For instance, red bricks are used in both building facades and sidewalk pavings (although different types of bricks are used for each purpose, they possess very close visual characteristics). Our goal is to have a model that can detect \emph{sidewalks} of certain materials from street-view images. \removed{so if we have similar textures on different objects, the model can distinguish between them.} 

Another challenging aspect of this task is the high degree of within-class variation and between-class similarities. For instance, NYC designated five different types of concrete as standard materials for sidewalk pavings, while Boston uses three different types of concrete. Each of these types has distinct visual features that, in some cases, can resemble materials of other classes, which pose further challenges to the classification task. Distinguishing between dark concrete and bluestone in some cases is very difficult, even for humans. When wet, some concretes with aggregates can look very similar to granite, and under the shadow, asphalt and worn-off concrete can look very similar.
Having a model that can accurately handle the within-class variability with between-class similarity calls for smartly selected training datasets with a good distribution of different classes as well as multiple variants of the same material under different conditions. 

\subsection{Limitations} 

Even though CitySurfaces can provide city-scale sidewalk material classification, some challenges remain unaddressed.
For instance, in the absence of proper sidewalk network data, it can be challenging to map the materials to their corresponding locations accurately. The maps in Figure~\ref{fig:overview} are based on the road centerlines where GSV cars traveled to capture images, depicting the dominant materials for each street segment by taking an average over the materials observed in each image from both the left and right sides of the street. 
However, knowing the exact location of certain materials is critical for urban designers, planners, and those working with safety and ease of walk for people with special needs. Although our model produces this result at a highly fine level, we cannot depict this variety in detail without proper sidewalk network data. Having separate maps for left and right sidewalks can be one solution, but the intersections where more than one street is captured pose a challenge for assigning the correct materials to each segment. 

Also, street-level images have some inherent limitations. Since the images are taken by cars moving alongside streets, in many instances, specifically in dense urban areas, the cars parked on the sides blocked the sidewalk view, as shown in the first street-view image of Figure~\ref{fig:gsv-diverse}. The issue can be mitigated to some extent by adjusting the heading and pitch of the camera, but that solution fails in images with large vehicles like trucks, or when the car \highlight{with mounted cameras} is too close to the sidewalks. 


\section{Conclusion}
\label{conclusion}
We present CitySurfaces, a scalable, low-cost approach towards the automatic computation of the spatial distribution of pavement materials at the sidewalk segment level. Our model can detect a diverse range of materials, which to our knowledge, were not covered by any existing dataset.
For instance, hexagonal pavers or granite blocks were not reported in any sidewalk inventories reviewed in this study. CitySurfaces produces accurate segmentation considering multiple cities both within and outside the domain of the training data, demonstrating generalization capabilities across varying urban fabrics. 
CitySurfaces can detect, delineate, and classify eight  standard surface materials used throughout most US cities. As shown in Section~\ref{block(c)}, the framework can be extended to include additional surface materials with less effort than building a city-specific model from scratch, which makes it possible for almost any city or government agency that has spatially dense street-level image data, to create a similar dataset.
Moreover, since we have covered the standard materials, such as concrete, asphalt, granite/bluestone, and brick, the model can be applied to a wide range of cities without any further annotation effort or with substantially less effort using our pre-trained model. \highlight{The models as well as the datasets generated for the six selected cities are publicly available in a GitHub repository.}

This work has addressed some challenges in data annotation and accurate classification of different materials with high between-class similarities and within-class variation. The active learning framework utilized in this study helped reduce the annotation costs by choosing the most informative set of data to be annotated and incrementally decreasing the manual modification time. 
By offering the first comprehensive dataset of sidewalk surface materials at the city scale, this study goes beyond reporting the dominant material of each segment and provides information on the percentage distribution of all detected materials per sidewalk segment. \removed{which combined with external data such as fine-scale shadow accumulation~\citep{shadowmap} can provide a more precise city-wide estimation of the surface UHI.} \highlight{The material classes in this study were selected based on the standard surface materials listed by Boston sidewalk inventory~\citep{Bostonswinv}, to use it as our baseline ground truth. That list is not extensive and does not distinguish between various types of the same class of material, such as concrete. However, for some more in-depth analysis, such as measuring UHI, we may need to classify the materials differently, and distinguish between different variations of the same material within one class. For instance, reflective granite and dark matte bluestone should have two distinct classes, same goes with the dark and light concretes since they have distinctively different albedo values. The CitySurfaces framework can be easily extended to detect more classes of materials as illustrated with the Manhattan example in Section~\ref{block(c)}, given the availability of the images corresponding to each class of interest to create the initial ground-truth set. In future work, we plan to take these differences into account and combine the generated data with shadow accumulation~\citep{shadowmap} to generate a city-scale UHI map.}

To facilitate designing automated audit tools, we are going to extend our model to detect surface problems such as potholes, significant breakage, and obstacles on pedestrian paths for accessibility analysis~\citep{miranda2020mosaic}. We also aim to address the walkability and active design of sidewalks by developing a model to detect the relevant features of the sidewalks wall plane and furnishing zone, such as window-to-wall ratio.
As another line for our future work, we would like to explore automated sample selection procedures and self-supervised learning techniques and tailor them to sidewalk and pedestrian facility analysis. 
We chose a simple (yet effective) uncertainty measure and coupled it with the analysis of the model's performance on the validation set and used expert's feedback to refine the annotations and check whether the model is predicting correctly since, on many instances, it is difficult to distinguish between visually similar materials.

\section*{Acknowledgment}

We would like to thank our colleagues at New York University and University of Illinois at Chicago for their help in this research.
This work was supported in part by: the Moore-Sloan Data Science Environment at NYU;
NASA; NSF awards CNS-1229185, CCF-1533564, CNS-1544753, CNS-1730396,
CNS-1828576, CNS-1626098; and the NVIDIA NVAIL at NYU. Claudio T. Silva is partially
supported by the DARPA D3M program. Any opinions, findings, and conclusions or
recommendations expressed in this material are those of the authors and do not necessarily
reflect the views of DARPA. We also thank Nvidia Corporation for donating GPUs used in this
research.

\appendix

\begin{figure}[t!]
\centering
\includegraphics[width=0.5\textwidth]{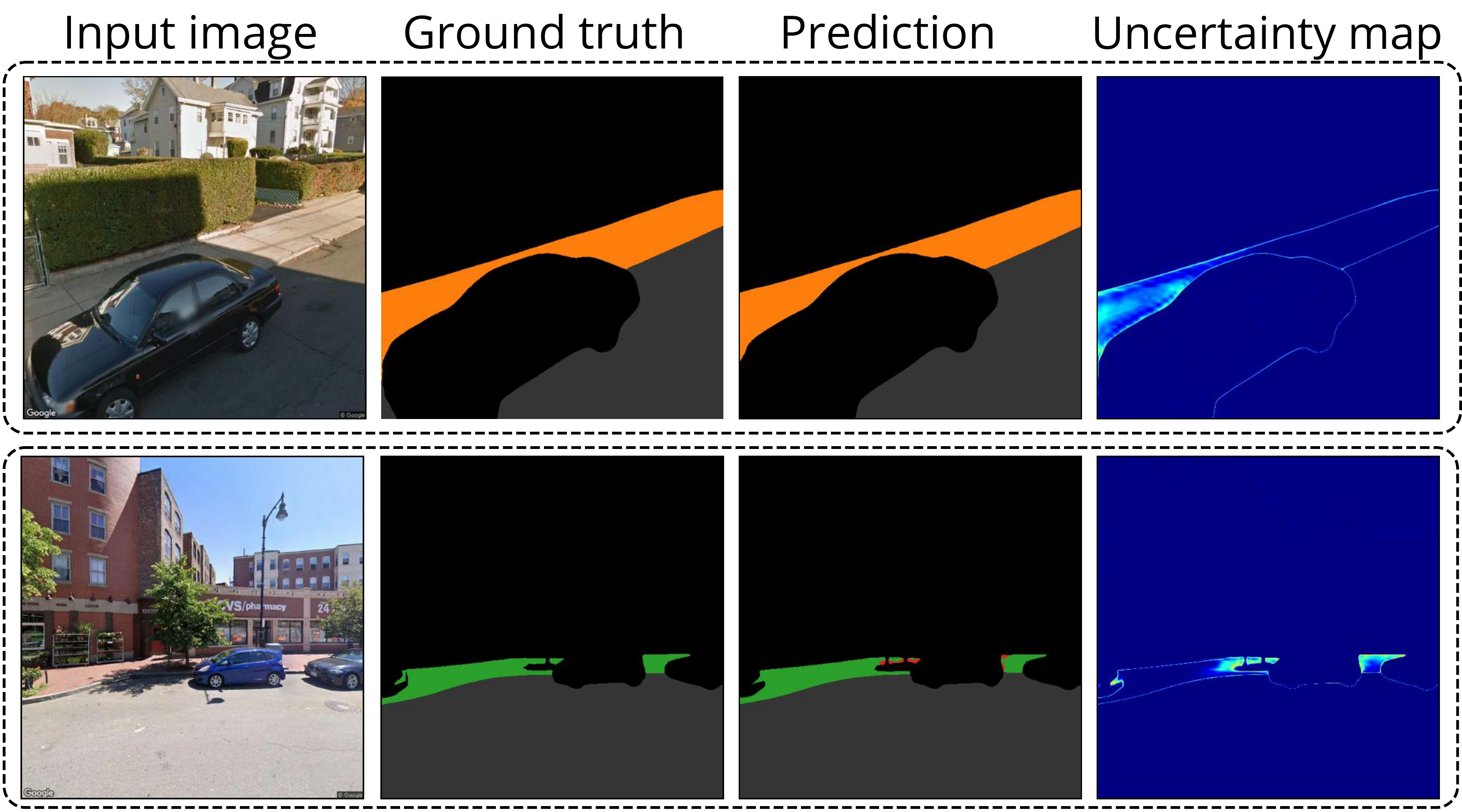}
\caption{Two different scenarios of using the model's output and uncertainty map in sample selection. The warmer colors in the uncertainty map represent areas where the model was less confident in its prediction. 
\textbf{Top}: the model correctly predicted the class in a previously identified challenging setting (shadow) but was less certain in predicting the shadowed areas; \textbf{Bottom}: The model classified the parts in shadow as concrete alongside brick and outputted mixed class for that region. The uncertainty map shows that the model was least certain in its prediction for that area.}
\label{fig:errorprob}
\end{figure}

\section{Sampling Strategies}
\label{appendixa}

\paragraph{i) Uncertainty in predicting unlabeled images}
Uncertainty sampling is one of the most frequently used query methods to select a new sample of training data in active learning \citep{settles2009active}. To measure the uncertainty, we use softmax probability, which has been commonly used in active learning as a strategy for choosing the next training sample~\citep{settles2009active}. We use the outputs of the softmax layer as part of the sampling strategy, which can partly reveal the most challenging instances for the model to predict. We apply multi-class uncertainty sampling known as margin sampling (MS)~\citep{scheffer2001active}, which calculates the difference between the two highest prediction probabilities on softmax to produce uncertainty maps. The smallest margin in each map is then chosen as the image-level uncertainty. The MS measure is defined as:
\begin{equation}
    x^*_{MS} = argmin_{x} P_{\theta}(\hat{y}_1|x) - P_{\theta}(\hat{y}_2|x)
\end{equation}
where $\hat{y}_1$ and $\hat{y}_2$ are the class labels for pixel $x$, with the first and second highest probability, respectively, under the model $\theta$. The lowest margin gives us the highest uncertainty, which is used as an image-level uncertainty measure. 

To select new samples, we feed the pool of unlabeled images to our network, obtain the segmentation and calculate image-level uncertainty to select images with the highest uncertainty. We start by selecting 10\% of the images using this strategy. As the training proceeds, we increase the share of images selected through this strategy at each stage by 10\%. 

\paragraph{ii) Performance on validation set}
Since softmax probabilities do not necessarily represent the true \emph{correctness} likelihood, a problem known as ``confidence calibration" \citep{guo2017calibration}, we need other strategies as well to select an informative sample for the model. To this end, at each stage, we examine the performance of the best epoch on the validation set and select 10\% of the best predictions and 20\% of the top failures. Images from failure and success cases are then clustered using K-means~\citep{cover1967nearest,fix1985discriminatory} with the Euclidean distance to investigate potential common patterns in each group. In each cluster, we rank images based on the average IoU of all classes, excluding road and background. We then select images with the highest error rate. The error rate is defined as the sums of false positive and false negative predictions of the model in each image. 
Aside from the described method, we examine the clusters of images to detect common error-causing patterns. Figure~\ref{fig:errorprob} (bottom row) depicts a brick sidewalk that the initial model incorrectly segmented the part next to shadowed regions as the ``mixed" class. Its associated uncertainty map reveals prediction difficulty near the edge of the car and the plant pit, which are incorrectly classified as mixed. Uncertainty maps of the success cases are examined to find regions where the model is least confident while making a correct prediction. Figure~\ref{fig:errorprob} highlights a set of uncertainty maps. After we find the most error-prone images, we use them to find similar unlabeled images. We extract their feature maps using the backbone HRNet-W48~\citep{sun2019high, wang2020deep} (more details in Section~\ref{hrnet}) and employ cosine similarity distance to retrieve similar images from the pool of unlabeled data.

\clearpage
\bibliographystyle{model5-names}\biboptions{authoryear}
\bibliography{paper}    

\begin{thebibliography}{66}
\expandafter\ifx\csname natexlab\endcsname\relax\def\natexlab#1{#1}\fi
\providecommand{\url}[1]{\texttt{#1}}
\providecommand{\href}[2]{#2}
\providecommand{\path}[1]{#1}
\providecommand{\DOIprefix}{doi:}
\providecommand{\ArXivprefix}{arXiv:}
\providecommand{\URLprefix}{URL: }
\providecommand{\Pubmedprefix}{pmid:}
\providecommand{\doi}[1]{\href{http://dx.doi.org/#1}{\path{#1}}}
\providecommand{\Pubmed}[1]{\href{pmid:#1}{\path{#1}}}
\providecommand{\bibinfo}[2]{#2}
\ifx\xfnm\relax \def\xfnm[#1]{\unskip,\space#1}\fi
\bibitem[{Agathangelidis et~al.(2020)Agathangelidis, Cartalis \&
  Santamouris}]{agathangelidis2020urban}
\bibinfo{author}{Agathangelidis, I.}, \bibinfo{author}{Cartalis, C.}, \&
  \bibinfo{author}{Santamouris, M.} (\bibinfo{year}{2020}).
\newblock \bibinfo{title}{Urban morphological controls on surface thermal
  dynamics: A comparative assessment of major european cities with a focus on
  athens, greece}.
\newblock {\it \bibinfo{journal}{Climate}\/},  {\it \bibinfo{volume}{8}\/},
  \bibinfo{pages}{131}.
\bibitem[{Aghaabbasi et~al.(2018)Aghaabbasi, Moeinaddini, Shah, Asadi-Shekari
  \& Kermani}]{AGHAABBASI2018475}
\bibinfo{author}{Aghaabbasi, M.}, \bibinfo{author}{Moeinaddini, M.},
  \bibinfo{author}{Shah, M.~Z.}, \bibinfo{author}{Asadi-Shekari, Z.}, \&
  \bibinfo{author}{Kermani, M.~A.} (\bibinfo{year}{2018}).
\newblock \bibinfo{title}{Evaluating the capability of walkability audit tools
  for assessing sidewalks}.
\newblock {\it \bibinfo{journal}{Sustainable Cities and Society}\/},  {\it
  \bibinfo{volume}{37}\/}, \bibinfo{pages}{475 -- 484}.
\bibitem[{Ai \& Tsai(2016)}]{ai2016automated}
\bibinfo{author}{Ai, C.}, \& \bibinfo{author}{Tsai, Y.} (\bibinfo{year}{2016}).
\newblock \bibinfo{title}{Automated sidewalk assessment method for americans
  with disabilities act compliance using three-dimensional mobile lidar}.
\newblock {\it \bibinfo{journal}{Transportation Research Record: Journal of the
  Transportation Research Board}\/},  (pp. \bibinfo{pages}{25--32}).
\bibitem[{Akbari et~al.(2009)Akbari, Menon \& Rosenfeld}]{akbari2009global}
\bibinfo{author}{Akbari, H.}, \bibinfo{author}{Menon, S.}, \&
  \bibinfo{author}{Rosenfeld, A.} (\bibinfo{year}{2009}).
\newblock \bibinfo{title}{Global cooling: increasing world-wide urban albedos
  to offset co 2}.
\newblock {\it \bibinfo{journal}{Climatic change}\/},  {\it
  \bibinfo{volume}{94}\/}, \bibinfo{pages}{275--286}.
\bibitem[{Akbari \& Rose(2008)}]{akbari2008urban}
\bibinfo{author}{Akbari, H.}, \& \bibinfo{author}{Rose, L.~S.}
  (\bibinfo{year}{2008}).
\newblock \bibinfo{title}{Urban surfaces and heat island mitigation
  potentials}.
\newblock {\it \bibinfo{journal}{Journal of the Human-environment System}\/},
  {\it \bibinfo{volume}{11}\/}, \bibinfo{pages}{85--101}.
\bibitem[{Amati \& Taylor(2010)}]{amati2010green}
\bibinfo{author}{Amati, M.}, \& \bibinfo{author}{Taylor, L.}
  (\bibinfo{year}{2010}).
\newblock \bibinfo{title}{From green belts to green infrastructure}.
\newblock {\it \bibinfo{journal}{Planning Practice \& Research}\/},  {\it
  \bibinfo{volume}{25}\/}, \bibinfo{pages}{143--155}.
\bibitem[{Anguelov et~al.(2010)Anguelov, Dulong, Filip, Frueh, Lafon, Lyon,
  Ogale, Vincent \& Weaver}]{anguelov2010google}
\bibinfo{author}{Anguelov, D.}, \bibinfo{author}{Dulong, C.},
  \bibinfo{author}{Filip, D.}, \bibinfo{author}{Frueh, C.},
  \bibinfo{author}{Lafon, S.}, \bibinfo{author}{Lyon, R.},
  \bibinfo{author}{Ogale, A.}, \bibinfo{author}{Vincent, L.}, \&
  \bibinfo{author}{Weaver, J.} (\bibinfo{year}{2010}).
\newblock \bibinfo{title}{{Google Street View}: Capturing the world at street
  level}.
\newblock {\it \bibinfo{journal}{Computer}\/},  {\it \bibinfo{volume}{43}\/},
  \bibinfo{pages}{32--38}.
\bibitem[{Arnold~Jr \& Gibbons(1996)}]{arnold1996impervious}
\bibinfo{author}{Arnold~Jr, C.~L.}, \& \bibinfo{author}{Gibbons, C.~J.}
  (\bibinfo{year}{1996}).
\newblock \bibinfo{title}{Impervious surface coverage: the emergence of a key
  environmental indicator}.
\newblock {\it \bibinfo{journal}{Journal of the American planning
  Association}\/},  {\it \bibinfo{volume}{62}\/}, \bibinfo{pages}{243--258}.
\bibitem[{Bell et~al.(2019)Bell, Tague \& McMillan}]{bell2019modeling}
\bibinfo{author}{Bell, C.~D.}, \bibinfo{author}{Tague, C.~L.}, \&
  \bibinfo{author}{McMillan, S.~K.} (\bibinfo{year}{2019}).
\newblock \bibinfo{title}{Modeling runoff and nitrogen loads from a watershed
  at different levels of impervious surface coverage and connectivity to storm
  water control measures}.
\newblock {\it \bibinfo{journal}{Water Resources Research}\/},  {\it
  \bibinfo{volume}{55}\/}, \bibinfo{pages}{2690--2707}.
\bibitem[{Bloodgood \& Vijay-Shanker(2014)}]{bloodgood2014method}
\bibinfo{author}{Bloodgood, M.}, \& \bibinfo{author}{Vijay-Shanker, K.}
  (\bibinfo{year}{2014}).
\newblock \bibinfo{title}{A method for stopping active learning based on
  stabilizing predictions and the need for user-adjustable stopping}.
\newblock {\it \bibinfo{journal}{arXiv preprint arXiv:1409.5165}\/}, .
\bibitem[{Boeing(2017)}]{boeing2017osmnx}
\bibinfo{author}{Boeing, G.} (\bibinfo{year}{2017}).
\newblock \bibinfo{title}{Osmnx: New methods for acquiring, constructing,
  analyzing, and visualizing complex street networks}.
\newblock {\it \bibinfo{journal}{Computers, Environment and Urban Systems}\/},
  {\it \bibinfo{volume}{65}\/}.
\bibitem[{{Boston PWD}(2014)}]{Bostonswinv}
\bibinfo{author}{{Boston PWD}} (\bibinfo{year}{2014}).
\newblock \bibinfo{title}{{Boston Sidewalk Inventory}}.
\newblock \URLprefix \url{https://data.boston.gov/dataset/sidewalk-inventory}.
\bibitem[{Chen et~al.(2016)Chen, Yang, Wang, Xu \& Yuille}]{chen2016attention}
\bibinfo{author}{Chen, L.-C.}, \bibinfo{author}{Yang, Y.},
  \bibinfo{author}{Wang, J.}, \bibinfo{author}{Xu, W.}, \&
  \bibinfo{author}{Yuille, A.~L.} (\bibinfo{year}{2016}).
\newblock \bibinfo{title}{Attention to scale: Scale-aware semantic image
  segmentation}.
\newblock In {\it \bibinfo{booktitle}{Proceedings of the IEEE Conference on
  Computer Vision and Pattern Recognition}\/} (pp.
  \bibinfo{pages}{3640--3649}).
\bibitem[{Chen \& Zhang(2017)}]{chen2017impacts}
\bibinfo{author}{Chen, X.}, \& \bibinfo{author}{Zhang, Y.}
  (\bibinfo{year}{2017}).
\newblock \bibinfo{title}{Impacts of urban surface characteristics on
  spatiotemporal pattern of land surface temperature in kunming of china}.
\newblock {\it \bibinfo{journal}{Sustainable Cities and Society}\/},  {\it
  \bibinfo{volume}{32}\/}, \bibinfo{pages}{87--99}.
\bibitem[{Chippendale \& Boltz(2015)}]{chippendale2015}
\bibinfo{author}{Chippendale, T.}, \& \bibinfo{author}{Boltz, M.}
  (\bibinfo{year}{2015}).
\newblock \bibinfo{title}{The neighborhood environment: perceived fall risk,
  resources, and strategies for fall prevention}.
\newblock {\it \bibinfo{journal}{The Gerontologist}\/},  {\it
  \bibinfo{volume}{55}\/}, \bibinfo{pages}{575--583}.
\bibitem[{Chithra et~al.(2015)Chithra, Nair, Amarnath \&
  Anjana}]{chithra2015impacts}
\bibinfo{author}{Chithra, S.}, \bibinfo{author}{Nair, M.~H.},
  \bibinfo{author}{Amarnath, A.}, \& \bibinfo{author}{Anjana, N.}
  (\bibinfo{year}{2015}).
\newblock \bibinfo{title}{Impacts of impervious surfaces on the environment}.
\newblock {\it \bibinfo{journal}{International Journal of Engineering Science
  Invention}\/},  {\it \bibinfo{volume}{4}\/}, \bibinfo{pages}{27--31}.
\bibitem[{Clifton et~al.(2007)Clifton, Smith \& Rodriguez}]{CLIFTON200795}
\bibinfo{author}{Clifton, K.~J.}, \bibinfo{author}{Smith, A. D.~L.}, \&
  \bibinfo{author}{Rodriguez, D.} (\bibinfo{year}{2007}).
\newblock \bibinfo{title}{The development and testing of an audit for the
  pedestrian environment}.
\newblock {\it \bibinfo{journal}{Landscape and Urban Planning}\/},  {\it
  \bibinfo{volume}{80}\/}, \bibinfo{pages}{95 -- 110}.
\bibitem[{Cordts et~al.(2016)Cordts, Omran, Ramos, Rehfeld, Enzweiler,
  Benenson, Franke, Roth \& Schiele}]{cordts2016cityscapes}
\bibinfo{author}{Cordts, M.}, \bibinfo{author}{Omran, M.},
  \bibinfo{author}{Ramos, S.}, \bibinfo{author}{Rehfeld, T.},
  \bibinfo{author}{Enzweiler, M.}, \bibinfo{author}{Benenson, R.},
  \bibinfo{author}{Franke, U.}, \bibinfo{author}{Roth, S.}, \&
  \bibinfo{author}{Schiele, B.} (\bibinfo{year}{2016}).
\newblock \bibinfo{title}{The cityscapes dataset for semantic urban scene
  understanding}.
\newblock In {\it \bibinfo{booktitle}{Proceedings of the IEEE Conference on
  Computer Vision and Pattern Recognition}\/} (pp.
  \bibinfo{pages}{3213--3223}).
\bibitem[{Cover \& Hart(1967)}]{cover1967nearest}
\bibinfo{author}{Cover, T.}, \& \bibinfo{author}{Hart, P.}
  (\bibinfo{year}{1967}).
\newblock \bibinfo{title}{Nearest neighbor pattern classification}.
\newblock {\it \bibinfo{journal}{IEEE Transactions on Information Theory}\/},
  {\it \bibinfo{volume}{13}\/}, \bibinfo{pages}{21--27}.
\bibitem[{Deitz et~al.(2021)Deitz, Lobben \& Alferez}]{deitz2021squeaky}
\bibinfo{author}{Deitz, S.}, \bibinfo{author}{Lobben, A.}, \&
  \bibinfo{author}{Alferez, A.} (\bibinfo{year}{2021}).
\newblock \bibinfo{title}{Squeaky wheels: Missing data, disability, and power
  in the smart city}.
\newblock {\it \bibinfo{journal}{Big Data \& Society}\/},  {\it
  \bibinfo{volume}{8}\/}.
\bibitem[{Du et~al.(2017)Du, Cai, Xu, Wang, Wang \& Cai}]{du2017quantifying}
\bibinfo{author}{Du, H.}, \bibinfo{author}{Cai, W.}, \bibinfo{author}{Xu, Y.},
  \bibinfo{author}{Wang, Z.}, \bibinfo{author}{Wang, Y.}, \&
  \bibinfo{author}{Cai, Y.} (\bibinfo{year}{2017}).
\newblock \bibinfo{title}{Quantifying the cool island effects of urban green
  spaces using remote sensing data}.
\newblock {\it \bibinfo{journal}{Urban Forestry \& Urban Greening}\/},  {\it
  \bibinfo{volume}{27}\/}, \bibinfo{pages}{24--31}.
\bibitem[{Estoque et~al.(2017)Estoque, Murayama \& Myint}]{estoque2017effects}
\bibinfo{author}{Estoque, R.~C.}, \bibinfo{author}{Murayama, Y.}, \&
  \bibinfo{author}{Myint, S.~W.} (\bibinfo{year}{2017}).
\newblock \bibinfo{title}{Effects of landscape composition and pattern on land
  surface temperature: An urban heat island study in the megacities of
  southeast asia}.
\newblock {\it \bibinfo{journal}{Science of the Total Environment}\/},  {\it
  \bibinfo{volume}{577}\/}, \bibinfo{pages}{349--359}.
\bibitem[{Ewing \& Handy(2009)}]{ewing2009measuring}
\bibinfo{author}{Ewing, R.}, \& \bibinfo{author}{Handy, S.}
  (\bibinfo{year}{2009}).
\newblock \bibinfo{title}{Measuring the unmeasurable: Urban design qualities
  related to walkability}.
\newblock {\it \bibinfo{journal}{Journal of Urban design}\/},  {\it
  \bibinfo{volume}{14}\/}, \bibinfo{pages}{65--84}.
\bibitem[{Fix(1985)}]{fix1985discriminatory}
\bibinfo{author}{Fix, E.} (\bibinfo{year}{1985}).
\newblock {\it \bibinfo{title}{Discriminatory analysis: nonparametric
  discrimination, consistency properties}\/} volume~\bibinfo{volume}{1}.
\newblock \bibinfo{publisher}{USAF School of Aviation Medicine}.
\bibitem[{Frackelton et~al.(2013)Frackelton, Grossman, Palinginis, Castrillon,
  Elango \& Guensler}]{frackelton2013measuring}
\bibinfo{author}{Frackelton, A.}, \bibinfo{author}{Grossman, A.},
  \bibinfo{author}{Palinginis, E.}, \bibinfo{author}{Castrillon, F.},
  \bibinfo{author}{Elango, V.}, \& \bibinfo{author}{Guensler, R.}
  (\bibinfo{year}{2013}).
\newblock \bibinfo{title}{Measuring walkability: Development of an automated
  sidewalk quality assessment tool}.
\newblock {\it \bibinfo{journal}{Suburban Sustainability}\/},  {\it
  \bibinfo{volume}{1}\/}, \bibinfo{pages}{4}.
\bibitem[{Glaeser et~al.(2018)Glaeser, Kominers, Luca \& Naik}]{glaeser2018big}
\bibinfo{author}{Glaeser, E.~L.}, \bibinfo{author}{Kominers, S.~D.},
  \bibinfo{author}{Luca, M.}, \& \bibinfo{author}{Naik, N.}
  (\bibinfo{year}{2018}).
\newblock \bibinfo{title}{Big data and big cities: The promises and limitations
  of improved measures of urban life}.
\newblock {\it \bibinfo{journal}{Economic Inquiry}\/},  {\it
  \bibinfo{volume}{56}\/}, \bibinfo{pages}{114--137}.
\bibitem[{Griew et~al.(2013)Griew, Hillsdon, Foster, Coombes, Jones \&
  Wilkinson}]{griew2013developing}
\bibinfo{author}{Griew, P.}, \bibinfo{author}{Hillsdon, M.},
  \bibinfo{author}{Foster, C.}, \bibinfo{author}{Coombes, E.},
  \bibinfo{author}{Jones, A.}, \& \bibinfo{author}{Wilkinson, P.}
  (\bibinfo{year}{2013}).
\newblock \bibinfo{title}{Developing and testing a street audit tool using
  google street view to measure environmental supportiveness for physical
  activity}.
\newblock {\it \bibinfo{journal}{International Journal of Behavioral Nutrition
  and Physical Activity}\/},  {\it \bibinfo{volume}{10}\/},
  \bibinfo{pages}{1--7}.
\bibitem[{Guo et~al.(2017)Guo, Pleiss, Sun \& Weinberger}]{guo2017calibration}
\bibinfo{author}{Guo, C.}, \bibinfo{author}{Pleiss, G.}, \bibinfo{author}{Sun,
  Y.}, \& \bibinfo{author}{Weinberger, K.~Q.} (\bibinfo{year}{2017}).
\newblock \bibinfo{title}{On calibration of modern neural networks}.
\newblock In {\it \bibinfo{booktitle}{International Conference on Machine
  Learning}\/} (pp. \bibinfo{pages}{1321--1330}).
\newblock \bibinfo{organization}{PMLR}.
\bibitem[{Huang et~al.(2010)Huang, Jin \& Zhou}]{huang2010active}
\bibinfo{author}{Huang, S.-J.}, \bibinfo{author}{Jin, R.}, \&
  \bibinfo{author}{Zhou, Z.-H.} (\bibinfo{year}{2010}).
\newblock \bibinfo{title}{Active learning by querying informative and
  representative examples}.
\newblock {\it \bibinfo{journal}{Advances in neural information processing
  systems}\/},  {\it \bibinfo{volume}{23}\/}, \bibinfo{pages}{892--900}.
\bibitem[{Jain \& Gruteser(2018)}]{jain2018recognizing}
\bibinfo{author}{Jain, S.}, \& \bibinfo{author}{Gruteser, M.}
  (\bibinfo{year}{2018}).
\newblock \bibinfo{title}{Recognizing textures with mobile cameras for
  pedestrian safety applications}.
\newblock {\it \bibinfo{journal}{IEEE Transactions on Mobile Computing}\/},
  {\it \bibinfo{volume}{18}\/}, \bibinfo{pages}{1911--1923}.
\bibitem[{Joshi et~al.(2021)Joshi, Leit{\~a}o, Maurer \& Bach}]{joshi2021not}
\bibinfo{author}{Joshi, P.}, \bibinfo{author}{Leit{\~a}o, J.~P.},
  \bibinfo{author}{Maurer, M.}, \& \bibinfo{author}{Bach, P.~M.}
  (\bibinfo{year}{2021}).
\newblock \bibinfo{title}{Not all suds are created equal: Impact of different
  approaches on combined sewer overflows}.
\newblock {\it \bibinfo{journal}{Water Research}\/},  {\it
  \bibinfo{volume}{191}\/}, \bibinfo{pages}{116780}.
\bibitem[{Kim et~al.(2020)Kim, Hwang, Lee, Kim, Choi \& Zhang}]{kim2020message}
\bibinfo{author}{Kim, T.}, \bibinfo{author}{Hwang, I.}, \bibinfo{author}{Lee,
  H.}, \bibinfo{author}{Kim, H.}, \bibinfo{author}{Choi, W.-S.}, \&
  \bibinfo{author}{Zhang, B.-T.} (\bibinfo{year}{2020}).
\newblock \bibinfo{title}{Message passing adaptive resonance theory for online
  active semi-supervised learning}.
\newblock {\it \bibinfo{journal}{arXiv preprint arXiv:2012.01227}\/}, .
\bibitem[{Kopf et~al.(2010)Kopf, Chen, Szeliski \& Cohen}]{kopf2010street}
\bibinfo{author}{Kopf, J.}, \bibinfo{author}{Chen, B.},
  \bibinfo{author}{Szeliski, R.}, \& \bibinfo{author}{Cohen, M.}
  (\bibinfo{year}{2010}).
\newblock \bibinfo{title}{Street slide: browsing street level imagery}.
\newblock (pp. \bibinfo{pages}{96:1--96:8}).
\newblock \bibinfo{organization}{ACM} volume~\bibinfo{volume}{29}.
\bibitem[{Lay et~al.(2020)Lay, Metcalf \& Sharp}]{lay2020paving}
\bibinfo{author}{Lay, M.}, \bibinfo{author}{Metcalf, J.}, \&
  \bibinfo{author}{Sharp, K.} (\bibinfo{year}{2020}).
\newblock {\it \bibinfo{title}{Paving Our Ways: A History of the World’s
  Roads and Pavements}\/}.
\newblock \bibinfo{publisher}{CRC Press}.
\bibitem[{Li et~al.(2013)Li, Zhou \& Ouyang}]{li2013relationship}
\bibinfo{author}{Li, X.}, \bibinfo{author}{Zhou, W.}, \&
  \bibinfo{author}{Ouyang, Z.} (\bibinfo{year}{2013}).
\newblock \bibinfo{title}{Relationship between land surface temperature and
  spatial pattern of greenspace: What are the effects of spatial resolution?}
\newblock {\it \bibinfo{journal}{Landscape and Urban Planning}\/},  {\it
  \bibinfo{volume}{114}\/}, \bibinfo{pages}{1--8}.
\bibitem[{{Loutzenheiser, Felix}(2010)}]{bostonregional}
\bibinfo{author}{{Loutzenheiser, Felix}} (\bibinfo{year}{2010}).
\newblock \bibinfo{title}{{Boston Region’s Pedestrian Transportation Plan}}.
\newblock \URLprefix
  \url{https://www.mapc.org/wp-content/uploads/2017/11/PedPlanFullReport.pdf}.
\bibitem[{Miranda et~al.(2019)Miranda, Doraiswamy, Lage, Wilson, Hsieh \&
  Silva}]{shadowmap}
\bibinfo{author}{Miranda, F.}, \bibinfo{author}{Doraiswamy, H.},
  \bibinfo{author}{Lage, M.}, \bibinfo{author}{Wilson, L.},
  \bibinfo{author}{Hsieh, M.}, \& \bibinfo{author}{Silva, C.~T.}
  (\bibinfo{year}{2019}).
\newblock \bibinfo{title}{Shadow accrual maps: Efficient accumulation of
  city-scale shadows over time}.
\newblock {\it \bibinfo{journal}{IEEE Transactions on Visualization and
  Computer Graphics}\/},  {\it \bibinfo{volume}{25}\/},
  \bibinfo{pages}{1559--1574}.
\bibitem[{Miranda et~al.(2020)Miranda, Hosseini, Lage, Doraiswamy, Dove \&
  Silva}]{miranda2020mosaic}
\bibinfo{author}{Miranda, F.}, \bibinfo{author}{Hosseini, M.},
  \bibinfo{author}{Lage, M.}, \bibinfo{author}{Doraiswamy, H.},
  \bibinfo{author}{Dove, G.}, \& \bibinfo{author}{Silva, C.~T.}
  (\bibinfo{year}{2020}).
\newblock \bibinfo{title}{Urban mosaic: Visual exploration of streetscapes
  using large-scale image data}.
\newblock In {\it \bibinfo{booktitle}{Proceedings of the 2020 CHI Conference on
  Human Factors in Computing Systems}\/} CHI ’20 (p.
  \bibinfo{pages}{1–15}).
\newblock \bibinfo{address}{New York, NY, USA}: \bibinfo{publisher}{Association
  for Computing Machinery}.
\bibitem[{Mooney et~al.(2016)Mooney, DiMaggio, Lovasi, Neckerman, Bader,
  Teitler, Sheehan, Jack \& Rundle}]{mooney2016use}
\bibinfo{author}{Mooney, S.~J.}, \bibinfo{author}{DiMaggio, C.~J.},
  \bibinfo{author}{Lovasi, G.~S.}, \bibinfo{author}{Neckerman, K.~M.},
  \bibinfo{author}{Bader, M.~D.}, \bibinfo{author}{Teitler, J.~O.},
  \bibinfo{author}{Sheehan, D.~M.}, \bibinfo{author}{Jack, D.~W.}, \&
  \bibinfo{author}{Rundle, A.~G.} (\bibinfo{year}{2016}).
\newblock \bibinfo{title}{Use of google street view to assess environmental
  contributions to pedestrian injury}.
\newblock {\it \bibinfo{journal}{American journal of public health}\/},  {\it
  \bibinfo{volume}{106}\/}, \bibinfo{pages}{462--469}.
\bibitem[{Muench et~al.(2010)Muench, Anderson \& Bevan}]{muench2010greenroads}
\bibinfo{author}{Muench, S.~T.}, \bibinfo{author}{Anderson, J.}, \&
  \bibinfo{author}{Bevan, T.} (\bibinfo{year}{2010}).
\newblock \bibinfo{title}{Greenroads: A sustainability rating system for
  roadways.}
\newblock {\it \bibinfo{journal}{International Journal of Pavement Research \&
  Technology}\/},  {\it \bibinfo{volume}{3}\/}.
\bibitem[{Nwakaire et~al.(2020)Nwakaire, Onn, Yap, Yuen \&
  Onodagu}]{nwakaire2020urban}
\bibinfo{author}{Nwakaire, C.~M.}, \bibinfo{author}{Onn, C.~C.},
  \bibinfo{author}{Yap, S.~P.}, \bibinfo{author}{Yuen, C.~W.}, \&
  \bibinfo{author}{Onodagu, P.~D.} (\bibinfo{year}{2020}).
\newblock \bibinfo{title}{Urban heat island studies with emphasis on urban
  pavements; a review}.
\newblock {\it \bibinfo{journal}{Sustainable Cities and Society}\/},  (p.
  \bibinfo{pages}{102476}).
\bibitem[{{NYC DOT}(2020)}]{new2020street}
\bibinfo{author}{{NYC DOT}} (\bibinfo{year}{2020}).
\newblock \bibinfo{title}{Street design manual}.
\bibitem[{Oke(1982)}]{oke1982energetic}
\bibinfo{author}{Oke, T.~R.} (\bibinfo{year}{1982}).
\newblock \bibinfo{title}{The energetic basis of the urban heat island}.
\newblock {\it \bibinfo{journal}{Quarterly Journal of the Royal Meteorological
  Society}\/},  {\it \bibinfo{volume}{108}\/}, \bibinfo{pages}{1--24}.
\bibitem[{Santamouris(2013)}]{santamouris2013using}
\bibinfo{author}{Santamouris, M.} (\bibinfo{year}{2013}).
\newblock \bibinfo{title}{Using cool pavements as a mitigation strategy to
  fight urban heat island—a review of the actual developments}.
\newblock {\it \bibinfo{journal}{Renewable and Sustainable Energy Reviews}\/},
  {\it \bibinfo{volume}{26}\/}, \bibinfo{pages}{224--240}.
\bibitem[{Santamouris et~al.(2011)Santamouris, Synnefa \&
  Karlessi}]{santamouris2011using}
\bibinfo{author}{Santamouris, M.}, \bibinfo{author}{Synnefa, A.}, \&
  \bibinfo{author}{Karlessi, T.} (\bibinfo{year}{2011}).
\newblock \bibinfo{title}{Using advanced cool materials in the urban built
  environment to mitigate heat islands and improve thermal comfort conditions}.
\newblock {\it \bibinfo{journal}{Solar Energy}\/},  {\it
  \bibinfo{volume}{85}\/}, \bibinfo{pages}{3085--3102}.
\bibitem[{Scheffer et~al.(2001)Scheffer, Decomain \&
  Wrobel}]{scheffer2001active}
\bibinfo{author}{Scheffer, T.}, \bibinfo{author}{Decomain, C.}, \&
  \bibinfo{author}{Wrobel, S.} (\bibinfo{year}{2001}).
\newblock \bibinfo{title}{Active hidden markov models for information
  extraction}.
\newblock In {\it \bibinfo{booktitle}{International Symposium on Intelligent
  Data Analysis}\/} (pp. \bibinfo{pages}{309--318}).
\newblock \bibinfo{organization}{Springer}.
\bibitem[{Settles(2009)}]{settles2009active}
\bibinfo{author}{Settles, B.} (\bibinfo{year}{2009}).
\newblock \bibinfo{title}{Active learning literature survey}.
\newblock {\it \bibinfo{journal}{CS Technical Reports}\/}, .
\bibitem[{Shuster et~al.(2005)Shuster, Bonta, Thurston, Warnemuende \&
  Smith}]{shuster2005impacts}
\bibinfo{author}{Shuster, W.~D.}, \bibinfo{author}{Bonta, J.},
  \bibinfo{author}{Thurston, H.}, \bibinfo{author}{Warnemuende, E.}, \&
  \bibinfo{author}{Smith, D.} (\bibinfo{year}{2005}).
\newblock \bibinfo{title}{Impacts of impervious surface on watershed hydrology:
  A review}.
\newblock {\it \bibinfo{journal}{Urban Water Journal}\/},  {\it
  \bibinfo{volume}{2}\/}, \bibinfo{pages}{263--275}.
\bibitem[{Sun et~al.(2019)Sun, Zhao, Jiang, Cheng, Xiao, Liu, Mu, Wang, Liu \&
  Wang}]{sun2019high}
\bibinfo{author}{Sun, K.}, \bibinfo{author}{Zhao, Y.}, \bibinfo{author}{Jiang,
  B.}, \bibinfo{author}{Cheng, T.}, \bibinfo{author}{Xiao, B.},
  \bibinfo{author}{Liu, D.}, \bibinfo{author}{Mu, Y.}, \bibinfo{author}{Wang,
  X.}, \bibinfo{author}{Liu, W.}, \& \bibinfo{author}{Wang, J.}
  (\bibinfo{year}{2019}).
\newblock \bibinfo{title}{High-resolution representations for labeling pixels
  and regions}.
\newblock {\it \bibinfo{journal}{arXiv preprint arXiv:1904.04514}\/}, .
\bibitem[{Takebayashi \& Moriyama()}]{polacco_study_2012}
\bibinfo{author}{Takebayashi, H.}, \& \bibinfo{author}{Moriyama, M.} ().
\newblock \bibinfo{title}{Study on surface heat budget of various pavements for
  urban heat island mitigation}, .
\newblock {\it \bibinfo{volume}{2012}\/}.
\newblock \bibinfo{note}{Publisher: Hindawi Publishing Corporation}.
\bibitem[{Talbot et~al.(2005)Talbot, Musiol, Witham \&
  Metter}]{talbot2005falls}
\bibinfo{author}{Talbot, L.~A.}, \bibinfo{author}{Musiol, R.~J.},
  \bibinfo{author}{Witham, E.~K.}, \& \bibinfo{author}{Metter, E.~J.}
  (\bibinfo{year}{2005}).
\newblock \bibinfo{title}{Falls in young, middle-aged and older community
  dwelling adults: perceived cause, environmental factors and injury}.
\newblock {\it \bibinfo{journal}{BMC public health}\/},  {\it
  \bibinfo{volume}{5}\/}, \bibinfo{pages}{1--9}.
\bibitem[{Tao et~al.(2020)Tao, Sapra \& Catanzaro}]{tao2020hierarchical}
\bibinfo{author}{Tao, A.}, \bibinfo{author}{Sapra, K.}, \&
  \bibinfo{author}{Catanzaro, B.} (\bibinfo{year}{2020}).
\newblock \bibinfo{title}{Hierarchical multi-scale attention for semantic
  segmentation}.
\newblock {\it \bibinfo{journal}{arXiv preprint arXiv:2005.10821}\/}, .
\bibitem[{Thomas et~al.(2020)Thomas, Gardiner, Crompton \&
  Lawson}]{thomas2020keep}
\bibinfo{author}{Thomas, N.~D.}, \bibinfo{author}{Gardiner, J.~D.},
  \bibinfo{author}{Crompton, R.~H.}, \& \bibinfo{author}{Lawson, R.}
  (\bibinfo{year}{2020}).
\newblock \bibinfo{title}{Keep your head down: Maintaining gait stability in
  challenging conditions}.
\newblock {\it \bibinfo{journal}{Human movement science}\/},  {\it
  \bibinfo{volume}{73}\/}, \bibinfo{pages}{102676}.
\bibitem[{Thomas M.~Menino(2013)}]{bostoncomplete}
\bibinfo{author}{Thomas M.~Menino, T. J.~T.} (\bibinfo{year}{2013}).
\newblock \bibinfo{title}{{Boston Complete Streets}}.
\newblock \URLprefix
  \url{https://tooledesign.com/project/boston-complete-streets-manual}.
\bibitem[{Tillson(1900)}]{tillson1900street}
\bibinfo{author}{Tillson, G.~W.} (\bibinfo{year}{1900}).
\newblock {\it \bibinfo{title}{Street Pavements and Paving Materials: A Manual
  of City Pavements: the Methods and Materials of Their Construction}\/}.
\newblock \bibinfo{publisher}{John Wiley \& Sons}.
\bibitem[{Van~Dam et~al.(2015)Van~Dam, Harvey, Muench, Smith, Snyder, Al-Qadi,
  Ozer, Meijer, Ram, Roesler et~al.}]{van2015towards}
\bibinfo{author}{Van~Dam, T.~J.}, \bibinfo{author}{Harvey, J.},
  \bibinfo{author}{Muench, S.~T.}, \bibinfo{author}{Smith, K.~D.},
  \bibinfo{author}{Snyder, M.~B.}, \bibinfo{author}{Al-Qadi, I.~L.},
  \bibinfo{author}{Ozer, H.}, \bibinfo{author}{Meijer, J.},
  \bibinfo{author}{Ram, P.}, \bibinfo{author}{Roesler, J.~R.} et~al.
  (\bibinfo{year}{2015}).
\newblock {\it \bibinfo{title}{Towards sustainable pavement systems: a
  reference document}\/}.
\newblock \bibinfo{type}{Technical Report} United States. Federal Highway
  Administration.
\bibitem[{Wang et~al.(2020)Wang, Sun, Cheng, Jiang, Deng, Zhao, Liu, Mu, Tan,
  Wang et~al.}]{wang2020deep}
\bibinfo{author}{Wang, J.}, \bibinfo{author}{Sun, K.}, \bibinfo{author}{Cheng,
  T.}, \bibinfo{author}{Jiang, B.}, \bibinfo{author}{Deng, C.},
  \bibinfo{author}{Zhao, Y.}, \bibinfo{author}{Liu, D.}, \bibinfo{author}{Mu,
  Y.}, \bibinfo{author}{Tan, M.}, \bibinfo{author}{Wang, X.} et~al.
  (\bibinfo{year}{2020}).
\newblock \bibinfo{title}{Deep high-resolution representation learning for
  visual recognition}.
\newblock {\it \bibinfo{journal}{IEEE Transactions on Pattern Analysis and
  Machine Intelligence}\/}, .
\bibitem[{Wu et~al.(2018)Wu, Sun, Li \& Yu}]{wu2018characterizing}
\bibinfo{author}{Wu, H.}, \bibinfo{author}{Sun, B.}, \bibinfo{author}{Li, Z.},
  \& \bibinfo{author}{Yu, J.} (\bibinfo{year}{2018}).
\newblock \bibinfo{title}{Characterizing thermal behaviors of various pavement
  materials and their thermal impacts on ambient environment}.
\newblock {\it \bibinfo{journal}{Journal of cleaner production}\/},  {\it
  \bibinfo{volume}{172}\/}, \bibinfo{pages}{1358--1367}.
\bibitem[{Yang et~al.(2019)Yang, Jin, Xiao, Jin, Xia, Li \&
  Wang}]{yang2019local}
\bibinfo{author}{Yang, J.}, \bibinfo{author}{Jin, S.}, \bibinfo{author}{Xiao,
  X.}, \bibinfo{author}{Jin, C.}, \bibinfo{author}{Xia, J.~C.},
  \bibinfo{author}{Li, X.}, \& \bibinfo{author}{Wang, S.}
  (\bibinfo{year}{2019}).
\newblock \bibinfo{title}{Local climate zone ventilation and urban land surface
  temperatures: Towards a performance-based and wind-sensitive planning
  proposal in megacities}.
\newblock {\it \bibinfo{journal}{Sustainable Cities and Society}\/},  {\it
  \bibinfo{volume}{47}\/}, \bibinfo{pages}{101487}.
\bibitem[{Yin et~al.(2015)Yin, Cheng, Wang \& Shao}]{yin2015big}
\bibinfo{author}{Yin, L.}, \bibinfo{author}{Cheng, Q.}, \bibinfo{author}{Wang,
  Z.}, \& \bibinfo{author}{Shao, Z.} (\bibinfo{year}{2015}).
\newblock \bibinfo{title}{‘big data’ for pedestrian volume: Exploring the
  use of google street view images for pedestrian counts}.
\newblock {\it \bibinfo{journal}{Applied Geography}\/},  {\it
  \bibinfo{volume}{63}\/}, \bibinfo{pages}{337--345}.
\bibitem[{Yuan et~al.(2019)Yuan, Chen \& Wang}]{yuan2019object}
\bibinfo{author}{Yuan, Y.}, \bibinfo{author}{Chen, X.}, \&
  \bibinfo{author}{Wang, J.} (\bibinfo{year}{2019}).
\newblock \bibinfo{title}{Object-contextual representations for semantic
  segmentation}.
\newblock {\it \bibinfo{journal}{arXiv preprint arXiv:1909.11065}\/}, .
\bibitem[{Zhang et~al.(2009)Zhang, Odeh \& Han}]{zhang2009bi}
\bibinfo{author}{Zhang, Y.}, \bibinfo{author}{Odeh, I.~O.}, \&
  \bibinfo{author}{Han, C.} (\bibinfo{year}{2009}).
\newblock \bibinfo{title}{Bi-temporal characterization of land surface
  temperature in relation to impervious surface area, ndvi and ndbi, using a
  sub-pixel image analysis}.
\newblock {\it \bibinfo{journal}{International Journal of Applied Earth
  Observation and Geoinformation}\/},  {\it \bibinfo{volume}{11}\/},
  \bibinfo{pages}{256--264}.
\bibitem[{Zhao et~al.(2019)Zhao, Wang, Yang \& Cai}]{zhao2019region}
\bibinfo{author}{Zhao, S.}, \bibinfo{author}{Wang, Y.}, \bibinfo{author}{Yang,
  Z.}, \& \bibinfo{author}{Cai, D.} (\bibinfo{year}{2019}).
\newblock \bibinfo{title}{Region mutual information loss for semantic
  segmentation}.
\newblock {\it \bibinfo{journal}{arXiv preprint arXiv:1910.12037}\/}, .
\bibitem[{Zhou et~al.(2017)Zhou, Zhao, Puig, Fidler, Barriuso \&
  Torralba}]{zhou2017scene}
\bibinfo{author}{Zhou, B.}, \bibinfo{author}{Zhao, H.}, \bibinfo{author}{Puig,
  X.}, \bibinfo{author}{Fidler, S.}, \bibinfo{author}{Barriuso, A.}, \&
  \bibinfo{author}{Torralba, A.} (\bibinfo{year}{2017}).
\newblock \bibinfo{title}{Scene parsing through ade20k dataset}.
\newblock In {\it \bibinfo{booktitle}{Proceedings of the IEEE Conference on
  Computer Vision and Pattern Recognition}\/} (pp. \bibinfo{pages}{633--641}).
\bibitem[{Zhu \& Mai(2019)}]{zhu2019review}
\bibinfo{author}{Zhu, S.}, \& \bibinfo{author}{Mai, X.} (\bibinfo{year}{2019}).
\newblock \bibinfo{title}{A review of using reflective pavement materials as
  mitigation tactics to counter the effects of urban heat island}.
\newblock {\it \bibinfo{journal}{Advanced Composites and Hybrid Materials}\/},
  {\it \bibinfo{volume}{2}\/}, \bibinfo{pages}{381--388}.
\bibitem[{Zhu et~al.(2019)Zhu, Sapra, Reda, Shih, Newsam, Tao \&
  Catanzaro}]{zhu2019improving}
\bibinfo{author}{Zhu, Y.}, \bibinfo{author}{Sapra, K.}, \bibinfo{author}{Reda,
  F.~A.}, \bibinfo{author}{Shih, K.~J.}, \bibinfo{author}{Newsam, S.},
  \bibinfo{author}{Tao, A.}, \& \bibinfo{author}{Catanzaro, B.}
  (\bibinfo{year}{2019}).
\newblock \bibinfo{title}{Improving semantic segmentation via video propagation
  and label relaxation}.
\newblock In {\it \bibinfo{booktitle}{Proceedings of the IEEE/CVF Conference on
  Computer Vision and Pattern Recognition}\/} (pp.
  \bibinfo{pages}{8856--8865}).

\end{thebibliography}
\end{document}